\documentclass[conference]{IEEEtran}
\usepackage{times}

\usepackage[numbers]{natbib}
\usepackage{multicol}
\usepackage[bookmarks=true,urlcolor=blue]{hyperref}
\usepackage{amsfonts}
\usepackage{amsmath}
\usepackage{booktabs}
\usepackage{graphicx}
\usepackage{gensymb}
\usepackage{color}
\usepackage{subcaption}
\newcommand{\dg}{\degree}
\newcommand{\tb}{\textbf}

\newcommand{\cD}{\mathcal{D}}

\newcommand{\cC}{\mathcal{C}}

\newcommand{\bX}{\mathbf{X}}
\newcommand{\bY}{\mathbf{Y}}
\newcommand{\bZ}{\mathbf{Z}}

\newcommand{\bYc}{\mathbf{Y}_{\mathcal{C}}}

\newcommand{\bI}{\mathbf{I}}

\newcommand{\bx}{\mathbf{x}}
\newcommand{\btx}{\mathbf{\tilde{x}}}

\newcommand{\by}{\mathbf{y}}
\newcommand{\bty}{\tilde{\mathbf{y}}}

\newcommand{\bp}{\mathbf{p}}
\newcommand{\btp}{\mathbf{ \tilde{p}}}

\newcommand{\bq}{\mathbf{q}}

\newcommand{\bW}{\mathbf{W}}

\newcommand{\bR}{\mathbf{R}}
\newcommand{\bt}{\mathbf{t}}

\newcommand{\bbR}{\mathbb{R}}

\begin{document}

\title{SASSE: Scalable and Adaptable 6-DOF Pose Estimation }

\author{Huu Le*, Tuan Hoang$^\dagger$,  Qianggong Zhang*, Thanh-Toan Do$^{\dagger\dagger}$, Anders Eriksson*, Michael Milford* \\
        \small{
        *Queensland University of Technology, 
        $^\dagger$Singapore University of Technology and Design,
        $^{\dagger\dagger}$University of Liverpool}

}









\maketitle

\begin{abstract}
Visual localization has become a key enabling component of many place recognition and SLAM systems. Contemporary research has primarily focused on improving accuracy and precision-recall type metrics, with relatively little attention paid to a system's absolute storage scaling characteristics, its flexibility to adapt to available computational resources, and its longevity with respect to easily incorporating newly learned or hand-crafted image descriptors. Most significantly, improvement in one of these aspects typically comes at the cost of others: for example, a snapshot-based system that achieves sub-linear storage cost typically provides no metric pose estimation, or, a highly accurate pose estimation technique is often ossified in adapting to recent advances in appearance-invariant features. In this paper, we present a novel 6-DOF localization system that for the first time \textit{simultaneously} achieves all the three characteristics: significantly sub-linear storage growth, agnosticism to image descriptors, and customizability to available storage and computational resources.  The key features of our method are developed based on a novel adaptation of multiple-label learning, together with effective dimensional reduction and learning techniques that enable simple and efficient optimization. We evaluate our system on several large benchmarking datasets and provide detailed comparisons to state-of-the-art systems. The proposed method demonstrates competitive accuracy with existing pose estimation methods while achieving better sub-linear storage scaling, significantly reduced absolute storage requirements, and faster training and deployment speeds. 

\end{abstract}

\IEEEpeerreviewmaketitle

\section{Introduction}
\label{sec:intro}

In recent years, the emergence of autonomous driving, robotics and augmented reality (AR) has relied in significant part on strong advances in the underlying localization and mapping algorithms~\cite{fuentes2015visual}. Several localization and mapping systems have achieved the ability to accurately operate in a wide variety of different visual conditions such as day-night, weather, and seasonal changes in the environment~\cite{netvlad}. By learning good representations for the scenes together with efficient encoding and hashing mechanisms, a few state-of-the-art localization systems are able to provide not only the rough location estimate, but also the accurate full six degree-of-freedom (6-DOF) camera poses~\cite{kendall2015posenet,brahmbhatt2018geometry,man-tip19}. 

\begin{figure}    
    \begin{subfigure}{\columnwidth}
        \centering
        \includegraphics[width=\columnwidth]{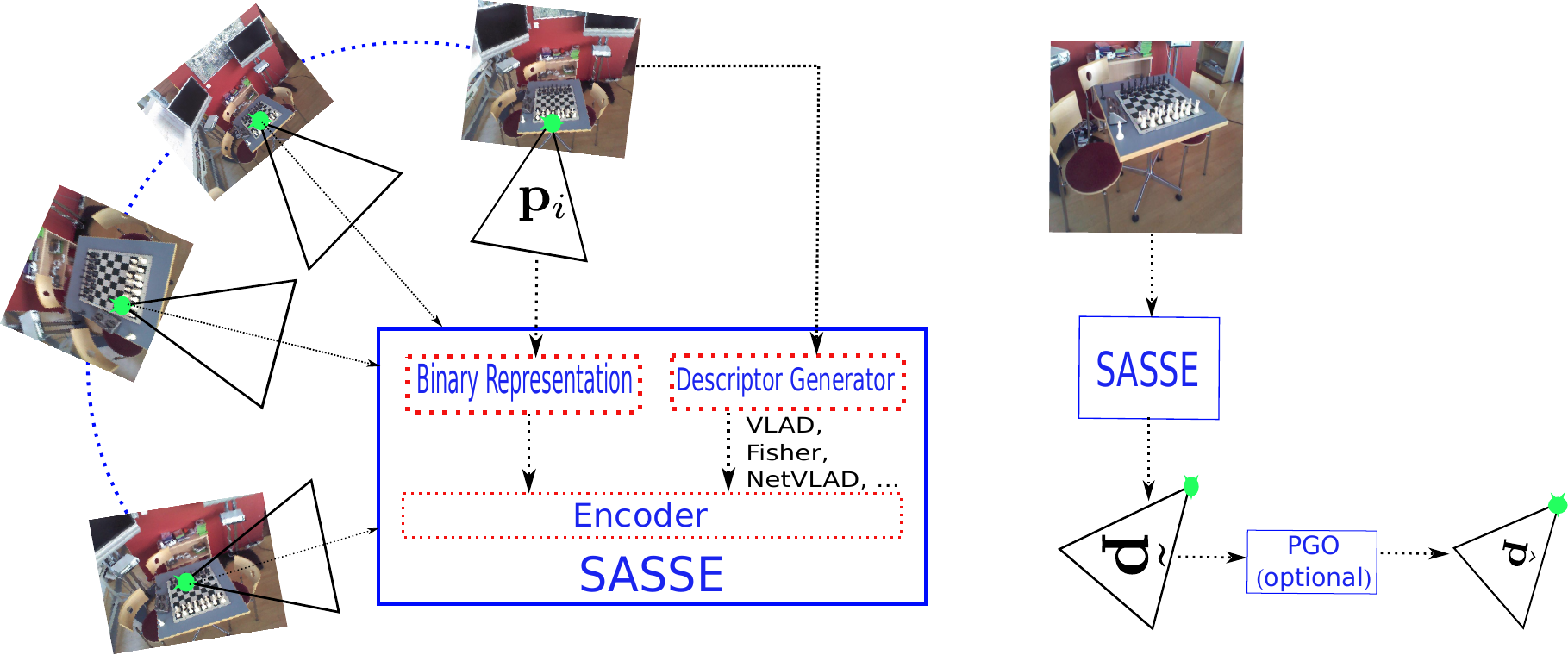}
        \caption{\textbf{Left: System design of SASSE.} RGB images and their corresponding poses are used for training. \textbf{Right: Inference pipeline.} From a single RBG image, SASSE predicts its camera location and orientation.} 
    \end{subfigure}

    \begin{subfigure}{\columnwidth}
        \centering
        \includegraphics[width=0.45\columnwidth]{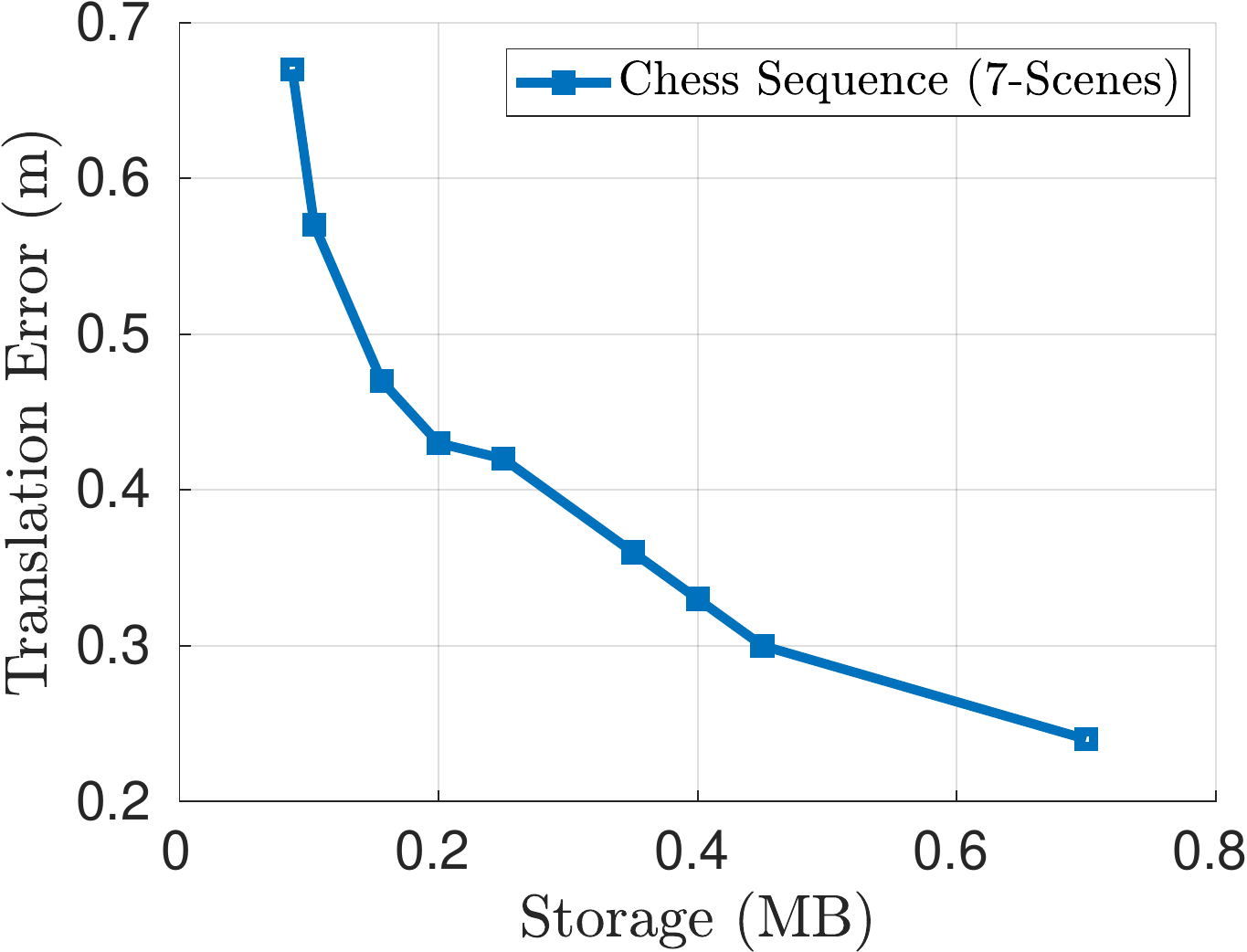}
        \includegraphics[width=0.45\columnwidth]{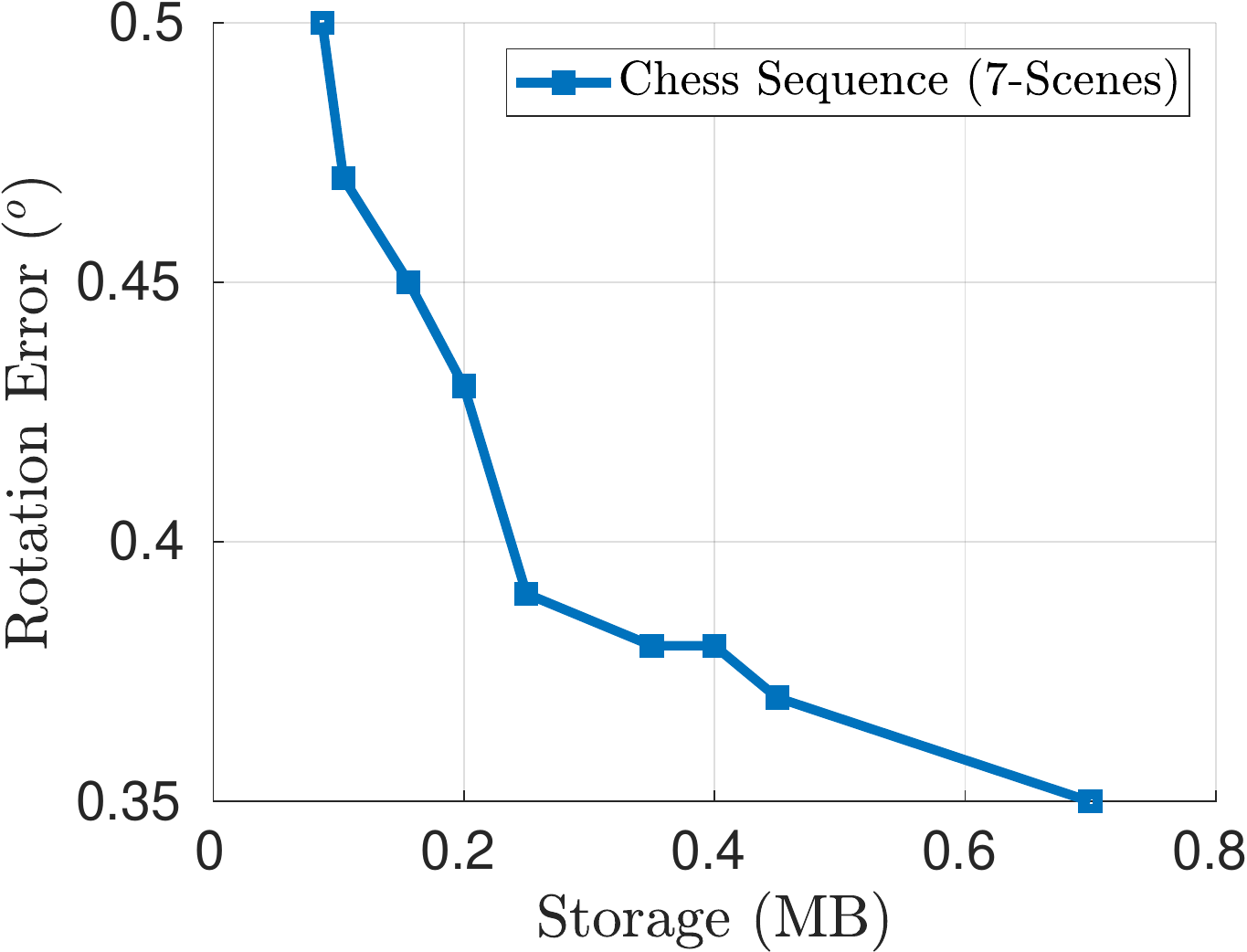}
        \caption{Performance of SASSE under the extreme case ($<$800KB of storage)  \tb{Left}: Translation error (m). \tb{Right}: Rotation error ($\dg$). SASSE can achieve an accurate localization of $<$$(0.25m, 0.36\degree)$ with only $0.7$MB of storage. }
    \end{subfigure}
    
    \caption{Overview of SASSE.  }
\end{figure}

While improving common metrics such as accuracy, precision and recall has been a primary focus of recent works on visual localization, the storage cost (and often correlated compute cost) of a system is also a key factor that needs to be carefully considered as the scale of training data grows. It becomes even more critical when a localization system is designed to deploy onto lightweight robots or AR devices with limited storage budgets. Additionally, in several long-term localization applications~\cite{hawes2017strands} where the input data constantly grow over time, the storage limits and the associated computational burdens become a bottleneck that impede the real-time performance of the overall system. Developing more efficient localization algorithms with less absolute storage consumption and better storage scalability is desirable because it offers a number of beneficial outcomes in general: more computational resources are spared for other system competencies, and the system can opt to cheaper, lighter, and less power hungry hardware. Within certain storage and computational budgets, a localization system has the options to either retaining the desired localization performance with less compute, or to use the saved resources to further improve localization accuracy. These benefits apply regardless of the absolute storage and computational resources available, whether it be a low power embedded system on a micro-drone or mini-robot, or a high-end autonomous driving system.

Although the precision and recall of existing localization algorithms have been significantly improved in recent years, their storage scalability has not been as thoroughly investigated. Our paper tackles the storage scalability issue by proposing a novel encoding algorithm that provides an accurate \emph{6-DOF localization} with significantly less storage footprint compared to existing approaches. Our method achieves significantly \emph{sub-linear} storage growth, i.e., its required storage scales significantly \emph{sub-linearly} with the size of the available input to maintain a certain degree of accuracy. Furthermore, by employing a suite of fast optimization techniques, the training and testing time of our algorithm is also substantially faster than existing approaches, which makes it suitable for many real-time applications.

The proposed approach also has longevity advantages compared with existing techniques. Despite the remarkable localization results achieved by existing pose regression methods such as PoseNet~\cite{kendall2015posenet} or MapNet~\cite{brahmbhatt2018geometry}, the ability of these networks to cope with extreme changes in visual conditions (e.g., day-night, weather, and seasons) is still less satisfactory, as empirically shown in~\cite{sattler2018benchmarking}. A large body of work has been devoted to improve the robustness to drastic environmental changes by developing effective global image descriptors that provide more informative representations of the scenes.  It would be very beneficial if the cutting-edge image descriptors can be utilized by the state-of-the-art localization methods for 6-DOF pose estimation. However, taking PoseNet and its variants~\cite{kendall2015posenet,kendall2017geometric} as examples, due to the fixed network structures, transferring and integrating of those advanced image descriptors (e.g., LoST~\cite{garg2018lost}) into the networks is not trivial, not to mention the fact that, for every type of feature descriptor, a new variant of the network needs to be trained and its hyper-parameters must be carefully tuned in order to achieve a good performance. Our proposed encoding mechanism does not have the same limitation.  It allows the flexible use of existing image descriptors, from the traditional hand-crafted features such as VLAD \cite{vlad} and Fisher Vector \cite{perronnin2007fisher} to recent state-of-the-art deep features such as NetVLAD \cite{netvlad}, DenseVLAD \cite{densevlad} and LoST~\cite{garg2018lost}. Hence, our method can effectively handle large environmental changes of the scenes while providing highly accurate 6-DOF localization results.

Our method also provides an easy mechanism for tuning the deployment size by  adjusting two hyper-parameters during the training process, allowing adjustment of the trade-off between storage and performance. In contrast, since the network structure of PoseNet and its variants are fixed, the configuration of their required storage is not easily modified.  Although recent network compression techniques \cite{dorefa,8099498} can be employed as a post-processing step to reduce the storage requirement, these techniques are usually complicated to use and require extra fine-tuning efforts for the compressed network to achieve comparable performance on the target training set.


\noindent \textbf{Contributions} In summary, the contributions of our work can be stated as follows:
\begin{itemize}
    \item We propose a novel storage-efficient encoding algorithm for 6-DOF localization (SASSE) using a novel adaptation of multiple-label learning techniques. The required storage of our system is \emph{order of magnitudes} lower than for existing methods such as PoseNet~\cite{kendall2015posenet} or MapNet~\cite{brahmbhatt2018geometry} while the localization results are comparable to these approaches.
    \item Our SASSE system allows the storage requirement to be easily tuned to suit the application at hand, enabling the system to be deployed onto a large variety of systems with different storage budgets. 
    \item The training and inference algorithms of SASSE are simple and easy to implement and the training and inference processes are fast. Our approach provides \emph{real-time performance} (approximately $7$ms/frame) without the need of GPUs, and is even faster than a GPU implementation of MapNet~\cite{brahmbhatt2018geometry}.
    \item Our system is agnostic to the input data, i.e., it is compatible with different types of image descriptors such as VLAD~\cite{vlad}, Fisher~\cite{perronnin2007fisher}, Democratic Embedding~\cite{jegou2014triangulation}, NetVLAD~\cite{netvlad}. To the best of our knowledge, we are the first to propose a method that allows state-of-the-art descriptors such as NetVLAD to be used for pose regression.
    \item We provide an empirical analysis on the scalability of our system, where we empirically show that SASSE achieves \emph{sub-linear storage growth} as the number of training frames increases.
\end{itemize}

Collectively, the contribution of this research is for the first time that all these desirable properties have been achieved simultaneously.
\section{Related Work}
Visual localization has become a mature research topic in recent years. 
A detailed survey can be referred to~\cite{lowry2016visual}. Generally speaking, existing methods for visual localization can be categorized into two classes of methods, namely, image retrieval based localization and accurate 6-DOF localization. 

The underlying mechanism behind retrieval-based localization methods is to learn a good image representation, which can then be used to measure the similarity between images. In the ideal case, the image of the same scene under different weather or lighting conditions should admit the same representation. In practice, the representations are not the same, however, we expect them to be as close as possible, so that the localization from a query image can be done by finding the nearest neighbors of its representation in the database. At its core, a global image descriptor can be generated by aggregating hand-crafted features, such as SIFT~\cite{lowe2004distinctive} or ORB~\cite{rublee2011orb}, to a global descriptor vector. The aggregating step can be done using the bag-of-word model, or the well-known  VLAD~\cite{vlad} and its variants~\cite{perronnin2007fisher,jegou2014triangulation}. Apart from conventional hand-crafted features, deep neural networks are also extensively utilized to obtain global image descriptors. Several good deep-network based descriptors have been proposed in the literature, such as NetVLAD~\cite{netvlad}, DenseVLAD~\cite{Torii-CVPR2015}. Recently, fine-tuning schemes~\cite{radenovic2018fine} have also been proposed which have achieved significant improvements in performance. 

On the other hand, 6-DOF localization methods aim to provide an accurate pose for a query image, including the camera location and its orientation with respect to the global environment. Traditional method builds the 3D point clouds for the scene using several structure-from-motion frameworks such as COLMAP~\cite{schonberger2016structure} or VisualSFM~\cite{wu2011visualsfm}. At the query stage, an image is localized by solving for the 2D-3D matching problem to register the image into the point cloud database~\cite{sattler2011fast}. In a wide operating environment, the number of 3D points can extremely large, thus finding the optimal 2D-3D matching is almost intractable. To mitigate this, several voting schemes~\cite{zeisl2015camera} have been developed. However, with badly contaminated data, the execution time of many random voting approaches may not be suitable for real-time localization.  Active Search~\cite{sattler2017efficient} has proposed a prioritized matching approach utilizing vocabulary trees to significantly speed up the query time.
To exploit the power of deep networks for 6-DOF localization, PoseNet proposes to train a network that directly regress the 6-DOF pose for a query image. During the training process, the image representations are implicitly learned, allowing PoseNet to accurately regress 6-DOF poses for new images in real time without explicitly performing 2D-3D matching. Several variants of PoseNet \cite{kendall2017geometric,walch2017image} and MapNet~\cite{brahmbhatt2018geometry} have been proposed since then and have been shown to achieve better results. The main drawback of deep-network based approaches is that the localization results are not as accurate as traditional 2D-3D matching frameworks, as shown in~\cite{sattler2018benchmarking}. Tree-based methods~\cite{shotton2013scene} have also been shown to perform well for several indoor datasets.

\section{Approach}
In this work, we assume that 6-DOF poses of the training images
are readily available. Given $N$ images represented by a set $\cD=\{(\bx_i, \bp_i)\}_{i=1}^N$, where $\bx_i \in \bbR^d$ is an image representation of image $i$, and $\bp_i$ denotes its corresponding pose. As previously mentioned, $\bx_i$ can be selected from a wide variety of global image descriptors~\cite{vlad,netvlad,finetune_hard_samples,perronnin2007fisher} and thus its dimensionality $d$ depends on the representation of choice. Our goal now is to learn a function $f(\cdot)$ such that, given a query image $\tilde{\bx}$, its pose $\tilde{\bp}$ can be best estimated by $f(\cdot)$, i.e., $\tilde{\bp} = f(\tilde{\bx})$.


\subsection{Pose Representation}
\label{sec:pose_representation}
Following other works on pose regression~\cite{kendall2015posenet,brahmbhatt2018geometry}, for each training image $\bx_i$, we represent its camera center by a three dimensional vector $\bt_i = \left[t^1_i, t_i^2, t_i^3\right] \in \bbR^3$ and its rotation by a quaternion $\bq_i$, which is a vector containing 4 real numbers: $\bq_i = \left[q_i^a,q_i^b,q_i^c,q_i^d\right]$. Thus, the pose $\bp_i$ of an image can be represented as a $7$-dimensional real vector:
\begin{equation}    
    \bp_i = \left[ \bq_i \;\; \bt_i^t\right] = \left[q_i^a,q_i^b,q_i^c,q_i^d,t^1_i, t_i^2, t_i^3\right]
    \label{eq:pose_description}
\end{equation}

\subsection{Binary Representation of a Pose Vector}
\label{sec:binary_representation}
Before training our encoder, we first transform the pose vectors described in~\eqref{eq:pose_description} into their corresponding binary representations. While several types of representation exist, we employ the simple approach of converting each element in the pose vector into a binary string, then concatenate the strings into one longer binary vector to represent the pose. More specifically, let $h(z)$ represent the operation that converts a real-valued number $z$ into its binary floating point representation. Thus, $h(z) \in \{0,1\}^b$, where the value of $b$ can be $16$, $32$, or $64$ depending on whether half, single, or double precision is used. For the ease of notation, we also denote $h(\bp)$ to be the operation that converts a pose vector $\bp$ into its binary representation:

\begin{equation}
    h(\bp_i) =\left[h(q_i^a) \; h(q_i^b)\; h(q_i^c) \; h(q_i^d) \; h(t^1_i) \; h(t_i^2) \; h(t_i^3)\right]
\end{equation}

It can be seen that $h(\bp_i)$ is a binary vector that has the length of $7b$, i.e., $h(\bp_i) \in \{0,1\}^{7b}$. The inverse operation of $h(\cdot)$, i.e., converting from a binary string into its original real-value representation, is denoted by $h^{-1}(\cdot)$.

We would like to remark that the reason for converting the pose vector $\bp_i$ into binary is for the user to explicitly choose the required encoding precision for the poses (by changing the value of $b$). As discussed in Section~\ref{sec:storage_analysis}, this will help to further reduce the absolute storage required. In our experiments, setting $b=16$ works well for most of the cases without any significant loss in accuracy compared to the experiments with $b=32$ or $b=64$.


\subsection{Binary Learning for Pose Regression}
\label{sec:training}
Henceforth, for brevity, we denote by $\bX \in \bbR^{N \times d}$ the matrix containing the training data $\{\bx_i\}_{i=1}^N$, where each row of $\bX$ represents a feature vector $\bx_i$. Similarly, let $\bY \in \{0,1\}^{N\times 7b}$ be the binary matrix that stores the poses of the training data, where each row of $\bY$ is a binary vector, i.e., $\by_i = h(\bp_i)$. The problem now becomes learning a function $f_b(\cdot) :\bbR^d \mapsto \{0,1\}^{7b}$ such that given a query vector $\tilde{\bx}$, the value of $h(\tilde{\bx})$ can be best predicted by $h(\tilde{\bx}) = f_b(\tilde{\bx})$. After obtaining $h(\tilde{\bx})$, its pose $\tilde{\bp}$ can be computed easily using the inverse operation $h^{-1}(\cdot)$. In other words, the predicted pose is $\tilde{\bp} = h^{-1}(f_b(\tilde{\bx}))$.

The overall learning algorithm of SASSE consists of two main steps, namely, learning the embedding of binary matrix and learning the regression in the reduced label space, which will be discussed in the following.

\subsubsection{Embedding of Label Matrix} \label{sec:label_embedding}

This step allows the binary label matrix $\bY$ to be embedded into a new domain with smaller dimensionality, thus it can also be referred to as the \emph{dimension reduction} step. As discussed in Section~\ref{sec:storage_analysis} and experimentally shown in Section~\ref{sec:embedding_size_tuning}, this step enables the required storage to be easily adjusted based on the applications' requirements. 

The main intuition behind this embedding step is to select a subset containing $r$ columns of $\bY$ ($r \le 7b$) such that the sub-matrix $\bYc \in \bbR^{N\times r}$ formed by the $r$ selected columns spans $\bY$ as much as possible. Mathematically, let $\cC \subseteq \{1,\dots,7b\}$ represent the set containing the indexes of the selected columns with $|\cC| = r$, the embedding step can be expressed as the following problem:
\begin{equation}
    \label{eq:label_embedding}
    \min_{\cC \subseteq \{1,\dots, 7b\}, |\cC| = r} \left\|\bY - \bYc \bYc^{\dagger}\bY\right\|_F,
\end{equation}
where the notation $^\dagger$ denotes the pseudo inverse of a matrix. The most straight-forward approach to solve~\eqref{eq:label_embedding} is to enumerate all possible subsets of size $r$ from the index set $\{1,\dots, 7b\}$ and return the subset that minimizes the objective of~\eqref{eq:label_embedding}. However, this is almost intractable when the number of columns in $\bY$ is large. To alleviate this difficulty, there exist several alternative approaches to solve~\eqref{eq:label_embedding} sub-optimally. In our work, we employ the method of exact subset sampling, which is inspired by~\cite{bi2013efficient}. More details are provided in the supplementary material.

After solving~\eqref{eq:label_embedding} to obtain the matrix $\bY_\cC$, we compute the projection matrix $\bZ$
\begin{equation}
\bZ = \bY_\cC^\dagger\bY.
\end{equation}
This matrix will be stored for the pose prediction process as it acts as a projection matrix from the reduced label space to the original space.  See details in Section~\ref{sec:pose_prediction}.
\subsubsection{Label Regression in the Reduced Space}\label{sec:ridge_regression}

After embedding the original binary label matrix into a smaller label space, i.e., every label vector $\by_i$ in the training dataset now corresponds to a reduced label vector $\by_{\cC i}$, the next step is to learn a regressor in this reduced label space. Particularly, we now aim to learn a function $f_r(\cdot): \bbR^d \mapsto \bbR^r$ such that given a query vector $\btx$, its reduced label $\bty_{\cC}$ is best predicted by $f_r(\cdot)$. Once $\bty_{\cC}$ is computed, we can project it back to the original label space then obtain the binary label vector $\bty$; details are discussed in Section~\ref{sec:pose_prediction}.

In this work, we propose to learn $f_r(\cdot)$ by solving a Ridge Regression problem, where the goal is to find a parameter matrix $\bW \in \bbR^{d\times r}$ that minimizes the following objective:
\begin{equation}
    \label{eq:ridge_regression}
    \min_{\bW \in \bbR^{d\times r}} \|\bX \bW  - \bYc\|^2_F + \lambda \|\bW\|^2_F,
\end{equation}
where $\lambda > 0$ is a regularization parameter. In our experiments, it is observed that the performance of SASSE is not sensitive to the parameter $\lambda$ so we fix $\lambda = 0.1$ throughout all the experiments.

One advantage our proposed learning scheme is that~\eqref{eq:ridge_regression} admits a closed-form solution $\bW^*$, which can be stated as:
\begin{equation}
    \label{eq:W_solution}
    \bW^* = \left(\bX^T\bX + \lambda\bI\right)^{-1}\bX^T \bY.
\end{equation}
Therefore, the learning of $f_r(\cdot)$ can be solved efficiently using matrix operations as expressed in~\eqref{eq:W_solution}.

\subsection{6-DOF Camera Pose Prediction}
\label{sec:pose_prediction}
It can now be seen that the task of learning the original pose predictor $f(\cdot)$ comprises two sub-tasks, namely, learning the projection matrix $\bZ = \bY_\cC^\dagger\bY$ and then learning the set of regression weights $\bW$ as discussed in Sections~\ref{sec:label_embedding} and~\ref{sec:ridge_regression}, respectively.
Therefore, given a query image represented by the feature vector $\btx \in \bbR^d$, its pose $\btp$ can be predicted by executing the following steps
\begin{itemize}
    \item Obtaining the reduced label $\bty_\cC \in \bbR^r$ using $f_r (\cdot)$
    \begin{equation}
        \bty_\cC = f_r(\bx) = \bW^T\btx 
    \end{equation}
    \item Projecting of $\bty_\cC$ back into the original binary label domain to obtain $\bty \in \{0,1\}^{7b}$:
    \begin{equation}
        \bty = \mathrm{sign}\left(\bty_{\cC}^T \bZ\right),
    \end{equation}
    where $\mathrm{sign}(a) = 1$ if $a>0$ and $0$ otherwise. We simply use the $\mathrm{sign}$ operation to obtain the binary label vector.
    \item Retrieving the 6-DOF pose: from the predicted vector $\bty$, the pose $\btp$ can be easily obtained by performing the inverse operation $h^{-1}(\cdot)$, i.e., converting the binary  vector $\bty$ to a real-valued vector containing the corresponding rotation and translation.
\end{itemize}

\subsection{Absolute Storage Analysis}
\label{sec:storage_analysis}
Based on our previous discussions, the parameters that needs to be stored include:

\begin{itemize}
    \item Regression matrix $\bW \in \bbR^{d\times r}$: as discussed in Section~\ref{sec:ridge_regression} and Section~\ref{sec:pose_prediction}, this matrix captures the mapping between the original feature space to the reduced label space. As can be observe from the dimension of $\bW$, its storage is independent of the database size $N$.
    \item Projection matrix $\bZ \in \bbR^{r\times 7b}$: this serves as the projection matrix that allows recovery of the original binary label from its corresponding reduced vector. Similar to $\bW$, the size of $\bZ$ does not depend on $N$.
\end{itemize}
From the analysis above, it is clear that in order to adjust the amount of storage required, one only needs to adjust the value of $r$, which is the dimension of the reduced label space, and $b$, which is the precision of the encoded poses. The total amount of required storage (in bytes) can then be expressed as:
\begin{equation}
    \label{eq:storage_no_cluster}
    S = 8(dr + 7rb) = 8r(d+7b),
\end{equation}
assuming that each real number requires $8$ bytes (64 bits). Refer to Section~\ref{sec:embedding_size_tuning} for experiments on how varying $r$ affects the absolute storage and the localization accuracy.

\subsection{Partition of Training Data}
Our proposed encoding algorithm is now functional for pose regression.  However, as a typical characteristic of many machine learning algorithms, when the number of training data grows drastically, this vanilla version is prone to under-fitting. To mitigate the under-fitting issue, inspired by the observation that the visual data for applications such as autonomous driving tends to be partitioned into different clusters - where image representations in each cluster are similar and can be easily classified, we propose to further advance the inference accuracy by introducing an additional clustering process. Particularly, we firstly partition the original training data into $k$ clusters $\cD_1, \dots \cD_k$, where $\cD_1\cup\dots\cup\cD_k = \cD$ and $\cD_i\cap\cD_j=\emptyset;\forall i \neq j$. Then, $k$ different pose regressors $\{f^i(\cdot)\}_{i=1}^k$, where $f^i(\cdot)$ takes $\cD_i$ as its training data, are trained using the procedure described in Section~\ref{sec:training}. 

At the inference stage, in order to predict the correct pose from a given query vector, the algorithm needs to select one of the $k$ possible predictors to use. This is achieved through an additional classifier $f_c(\cdot)$, where $f_c(\btx)$ provides the cluster index $j$ to which $\btx$ belongs, so that the corresponding predictor $f^j(\cdot)$ can then be employed to obtain the pose. We employ a Support Vector Machine (SVM) to train $f_c$, where the input data for the SVM is the original dataset $\cD$, while the training label of each data point is the index of the corresponding cluster to which it is assigned.

When clustering is applied, the absolute storage of SASSE must be computed as the total storage required by $k$ different regressors $f^i(\cdot)$, with the additional storage required to store the parameters of $f_c(\cdot)$. 
Specifically, as $f_c(\cdot)$ is a linear classifier with $k$ classes, $k-1$ hyperplanes need to be stored, and each hyperplane is of $d+1$ dimensions. Together with~\eqref{eq:storage_no_cluster}, the total required (in bytes) storage now becomes
\begin{equation}
    \label{eq:final_storage}
    S = 8\big(kr(d+7b) + (k-1)(d+1)\big),
\end{equation}
where $r$ is the embedding size as discussed in  Sec.~\ref{sec:label_embedding}, and $b$ is the precision described in Sec.~\ref{sec:binary_representation} 

From~\eqref{eq:final_storage}, it can be observed that besides $r$, the number of clusters $k$ is the additional parameter that can be tuned to achieve the desired storage. Since the dimension $d$ of the feature vector can also impact the storage, one can further reduce the storage by choosing a type of feature with smaller dimensionality. However, to demonstrate the impact of the two main hyper-parameters in our algorithm, we use the same value of $d$ throughout all experiments.

\subsection{Pose Correction with Pose Graph Optimization (PGO) }
We have explained the major techniques underpinning the SASSE, which demonstrates good localization results in the experiment section. Additionally, in some applications where relative poses among consecutive query frames are also made available, the absolute poses of each frames can be further refined by incorporated the relative poses into a PGO framework. 

Let us denote by $\hat{\bt}_i$ and $\hat{\bR}_i$ the predicted camera center and orientation of frame $i$, respectively. Assume that the relative poses between frames $i$ and $j$, denoted by $\bt_{ij}$ and $\bR_{ij}$,  is provided by an off-the-shelf Visual Odometry (VO) framework~\cite{engel2018direct}. The goal of PGO is to find the optimal poses $(\bt^*_i, \bR^*_i)$ and $(\bt^*_j,\bR^*_j) $ to minimize the following objective function~\cite{carlone2015initialization}:
\begin{equation}
    \label{eq:pose_optimization}
    \min_{\substack{\bt_i,\bt_j \in \bbR^3 \\ \bR_i, \bR_j \in SO(3)}} \sum_{\substack{i,j=1,\\ i\ne j}}^T L_{ij}  + \sum_{i=1}^T L'_i,
\end{equation}
where the functions $L_{ij}$ and $L'_i$ are defined as follows
\begin{equation*}    
    L_{ij} = \left\|\bt_{ij} - \bR^T_i (\bt_j - \bt_i)\right\|^2 + d^2_R\left(\bR_{ij}, \bR_i^T\bR_j\right)     
\end{equation*}
\begin{equation*}    
    L_{i} = \left\|\bt_{i} - \hat{\bt}_i\right\|^2 + d^2_R\left(\bR_{ij}, \bR_i^T\bR_j\right)     
\end{equation*}
Here, $d_R$ denotes the rotation error between the two rotation matrices and $T$ represents the number of frames in a window that needs to be optimized. In our experiments, PGO is executed for windows containing $T=5$ to $10$ frames.

While several standard PGO solvers~\cite{grisetti2010tutorial} can be employed to solve~\eqref{eq:pose_optimization}, our empirical results show that SASSE can produce very accurate orientation, where most of the rotation errors are less than $1\degree$. Therefore, we simplify the solving of~\eqref{eq:pose_optimization} by assuming that $\bR^*_i = \hat{\bR}_i$, i.e., the optimal poses are provided by SASSE. Thus, our PGO problem boils down to solving the convex least squares problem
\begin{equation}
    \label{eq:pgo_translation}
    \min_{\bt_i,\bt_j } \sum_{ij}\left\|\bt_j - \bt_i - \hat{\bR}_i\bt_{ij}\right\|^2 + \sum_i \left\|\bt_i - \hat{\bt}_i\right\|^2,
\end{equation}
which can be solved using a closed-form solution.

\section{Experimental Results}

\setlength{\tabcolsep}{2.7pt}
\begin{table*}[ht]
    \centering
    \resizebox{1.0\textwidth}{!}{
    \begin{tabular}{@{}ccc|ccccc||cc@{}}
    \toprule
    Scene   & \#Train & \#Test & PoseNet17~\cite{kendall2017geometric} & PoseNet-LSTM~\cite{walch2017image} & VidLoc~\cite{clark2017vidloc} & MapNet+PGO~\cite{brahmbhatt2018geometry} & SCoRe Forest~\cite{shotton2013scene} & SASSE & SASSE + PGO \\ \midrule
    Chess   & 4000  & 2000  &  0.13m, 4.48\dg & 0.24m, 5.77\dg  &  0.18m, NA  & 0.09m, 3.24\dg & \tb{0.03m}, 0.66\dg &  0.27m, \tb{0.41}\dg  & 0.19m, \tb{0.41}\dg  \\
    Fire    & 2000  & 2000  &  0.27m, 11.30\dg & 0.34m, 11.9\dg &  0.26m, NA  & 0.20m, 9.04\dg & \tb{0.05m}, 1.50\dg & 0.38m, \tb{0.37}\dg & 0.28m, \tb{0.37}\dg   \\    
    Heads   & 1000  & 1000  &  0.17m, 13.00\dg & 0.21m, 13.7\dg &  0.14m, NA  & 0.12m, 8.45\dg & \tb{0.06m}, 5.50\dg &  0.15m, \tb{0.32}\dg &0.11m, \tb{0.32}\dg   \\    
    Office  & 6000  & 4000  &  0.19m, 5.55\dg & 0.30m, 8.08\dg  &  0.26m, NA  & 0.17m, 5.15\dg & \tb{0.04m}, 0.78\dg &   0.38m, \tb{0.51}\dg & 0.38m, \tb{0.51}\dg  \\    
    Pumpkin & 4000  & 2000  &  0.26m, 4.75\dg & 0.33m, 7.0\dg &  0.36m, NA  & 0.22m, 4.02\dg & \tb{0.04m}, 0.68\dg &    0.38m, \tb{0.33}\dg & 0.32m, \tb{0.33}\dg\\    
    Kitchen & 7000  & 5000  &  0.23m, 5.35\dg & 0.37m, 8.83\dg &  0.31m, NA  & 0.20m, 4.93\dg & \tb{0.04m}, 0.76\dg &   0.44m, \tb{0.52}\dg &  0.31m, \tb{0.52}\dg \\    
    Stairs &  2000  & 1000  &  0.35m, 12.40\dg & 0.40m, 13.7\dg &  0.26m, NA  & 0.27m, 10.57\dg & \tb{0.32m}, 1.32\dg &  0.33m, \tb{0.51}\dg &  0.25m, \tb{0.34}\dg   \\    
    
    \bottomrule
    \end{tabular}
    }
    \caption{Experimental Results on the 7-Scenes Dataset. For SASSE, we provide both the results before and after PGO.  }
    \label{table:results_7scenes}
\end{table*}


\begin{table*}[ht]
    \centering
    \resizebox{1.0\textwidth}{!}{
    \begin{tabular}{@{}ccc|ccccc||cc@{}}
    \toprule
    Scene   & \#Train & \#Test & Active Search~\cite{sattler2017efficient} & PoseNet~\cite{kendall2015posenet} & Bayesian PoseNet~\cite{kendall2016modelling} & PoseNet-LSTM~\cite{walch2017image} & PoseNet17~\cite{kendall2017geometric} & SASSE & SASSE + PGO \\ \midrule
    
    Great Court   & 1533  & 761  &  -- & --  &  --  & -- & \tb{6.83m}, 3.47\dg &  18.3m, 0.53\dg   & 12.5m, \tb{0.53}\dg \\
    
    King's College    & 1220  & 343  &  \tb{0.42m}, 0.55\dg & 1.66m, 4.86\dg  &  1.74m, 4.06\dg  & 0.99m, 3.65\dg & 0.88m, 1.04\dg &  3.18m, \tb{0.14}\dg   &  1.61m, \tb{0.14}\dg\\
    
    Old Hospital   & 895  & 182  &  \tb{0.44m}, 1.01\dg & 2.62m, 4.90\dg  &  2.57m, 5.14\dg  & 1.51m, 4.29\dg & 3.20m, 3.29\dg &  4.18m, \tb{0.20}\dg   & 2.97m, \tb{0.20}\dg \\
    
    Shop Facade   & 231  & 103  &  \tb{0.12m}, 0.40\dg & 1.41m, 7.18\dg  &  1.25m, 7.54\dg  & 1.18m, 7.44\dg & 0.88m, 3.78\dg &  2.28m, \tb{0.21}\dg  & 1.59m \tb{0.21}\dg \\
    
    St Mary's Church    & 1487  & 530  &  \tb{0.19m}, 0.54\dg & 2.45m, 7.96\dg  &  2.11m, 8.38\dg  & 1.52m, 6.68\dg & 1.57m, 3.32\dg &  4.38m, \tb{0.23}\dg &  2.91m \tb{0.23}\dg \\
    
    Street   & 3015  & 2923  &  \tb{0.85m}, \tb{0.83}\dg & --  &  --  & -- & 20.3m, 25.5\dg &  26.61m, 1.11\dg   & 23.13m, 1.11\dg\\
    
    \bottomrule
    \end{tabular}
    }
    \caption{Experimental Results on the Cambridge Landmarks Dataset.}
    \label{table:results_Cambridge}
\end{table*}

\begin{table}[]
    \centering

    \resizebox{1.0\columnwidth}{!}{
        \begin{tabular}{@{}c|c|c|c|c|c@{}}
        \toprule
                & Chess            & Fire             & Heads            & Office            & Pumpkin            \\ \midrule
        SASSE & \textbf{9.8 MB} & \textbf{7.8 MB} & \textbf{7.5 MB} & \textbf{12.7 MB} & \textbf{15.9 MB}   \\ \midrule
        BTRF~\cite{meng2017backtracking}    &    130.5 MB      & 110.8 MB                 & 146.5 MB         &   144.8 MB        &    100.2 MB               \\ \bottomrule
        MapNet~\cite{brahmbhatt2018geometry}  & \multicolumn{5}{c}{256 MB}                                                                     \\   \midrule
        \end{tabular}        
    }
    \vspace{0.5cm}

    \resizebox{1.0\columnwidth}{!}{
        \begin{tabular}{@{}c|c|c|c|c|c@{}}
        \toprule
                & Great Court            & King's College             & Old Hospital            & Shop Facade            & Street            \\ \midrule
        SASSE & \textbf{14.4 MB} & \textbf{37.2 MB} & \textbf{12.9 MB} & \textbf{12.7 MB} & \textbf{15.9 MB}   \\ \midrule
        PoseNet-LSTM~\cite{walch2017image} & \multicolumn{5}{c}{64.3 MB}                                                                    \\  \midrule
        \end{tabular}
    }

    \caption{SASSE's absolute storage (in comparison to other methods) to achieve the localization results shown in Table~\ref{table:results_7scenes} and Table~\ref{table:results_Cambridge}. Top: 7-Scenes dataset;  Bottom: Cambridge Landmark dataset. Our method requires much smaller absolute storage, while the translation and rotation errors are comparable. Note that the storage for the VLAD's dictionary is also included.}
    \label{table:storage}
\end{table}

\begin{figure*}[h]
    \centering
    \begin{subfigure}{\textwidth}
        \centering
        \includegraphics[width = 0.3\textwidth]{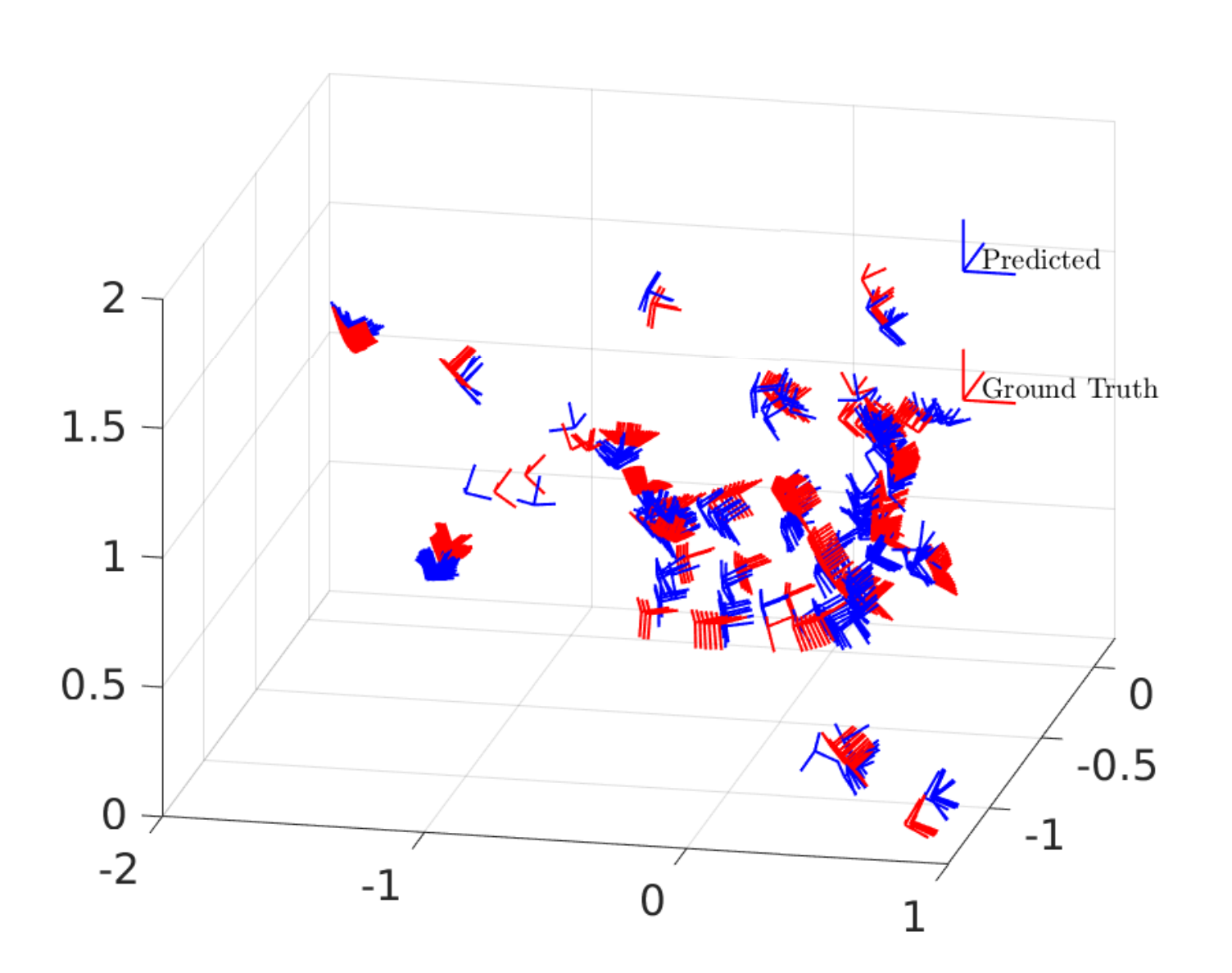}
        \includegraphics[width = 0.3\textwidth]{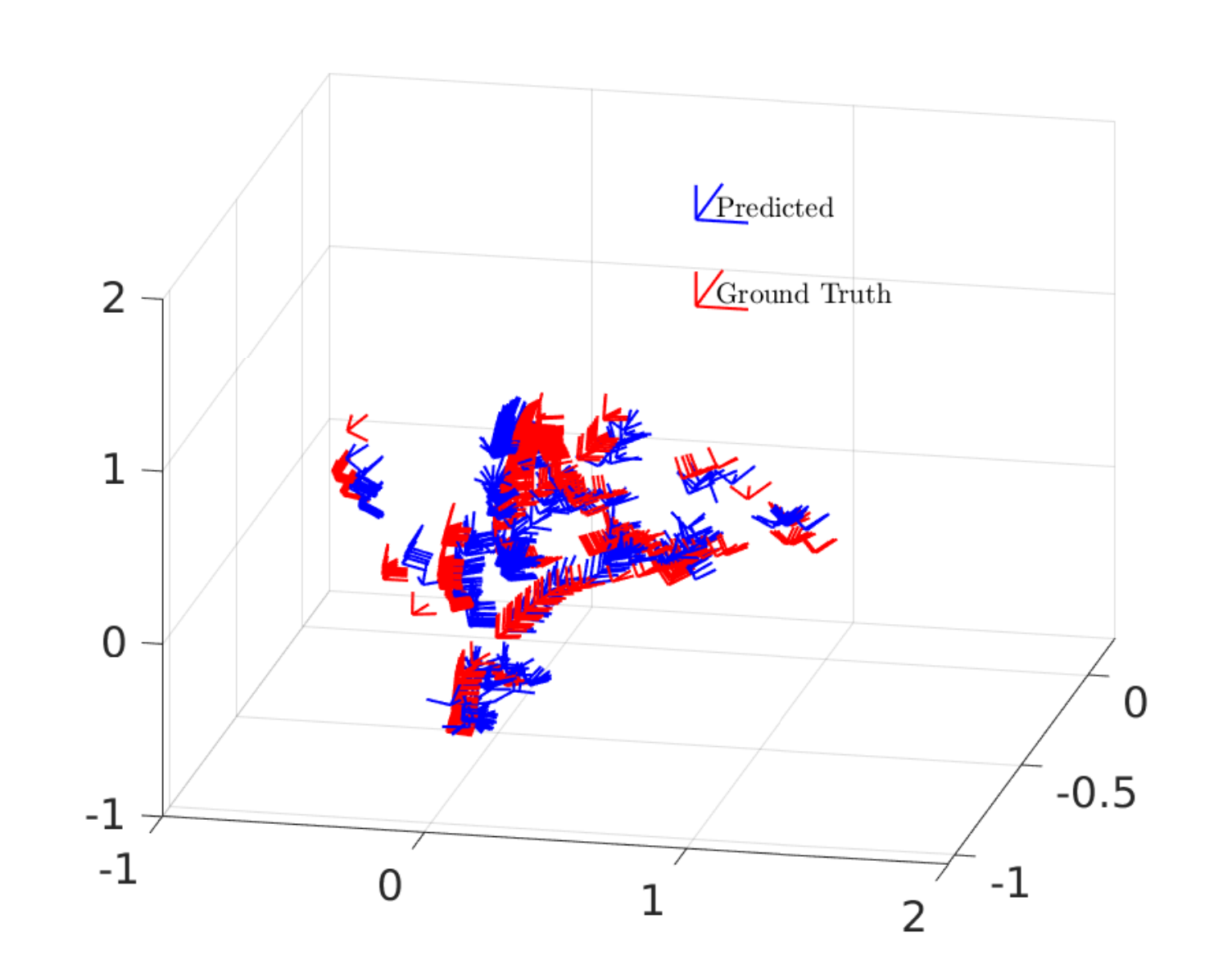}
        \includegraphics[width = 0.3\textwidth]{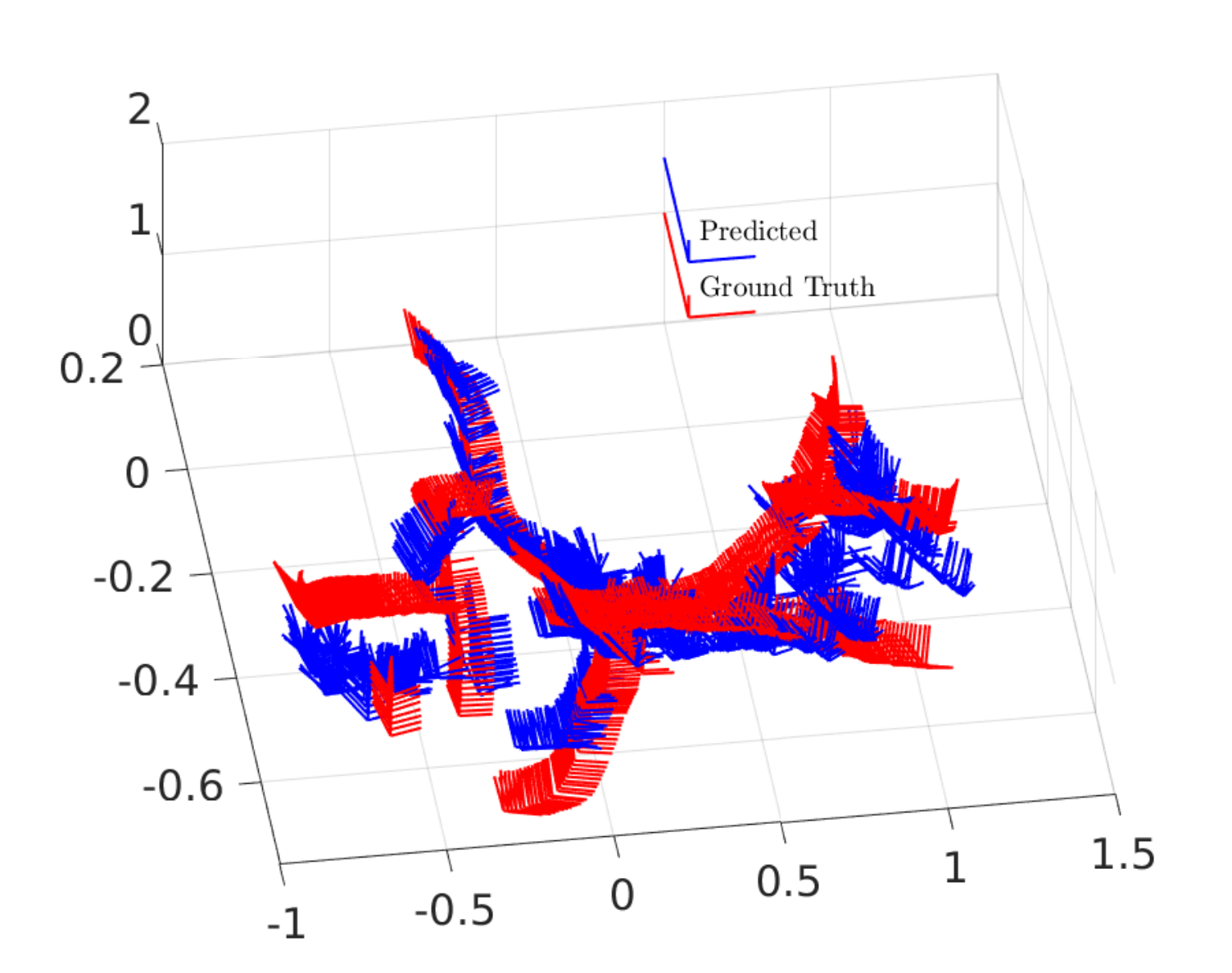}
        \caption{7-Scenes dataset. The sequences from left to right: Office, Chess, and Fire.}
    \end{subfigure}                

    \begin{subfigure}{\textwidth}
        \centering
        \includegraphics[width = 0.3\textwidth]{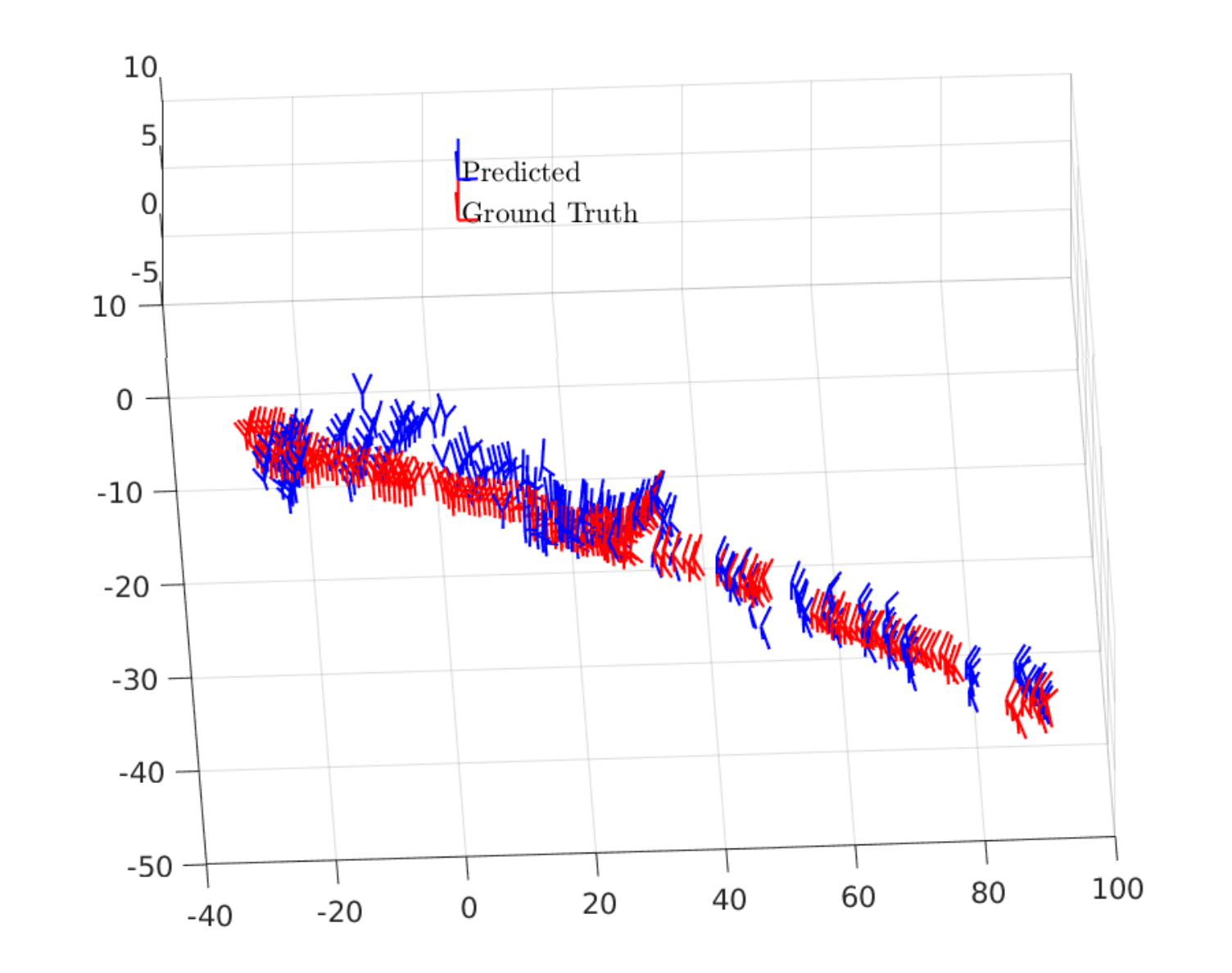}
        \includegraphics[width = 0.3\textwidth]{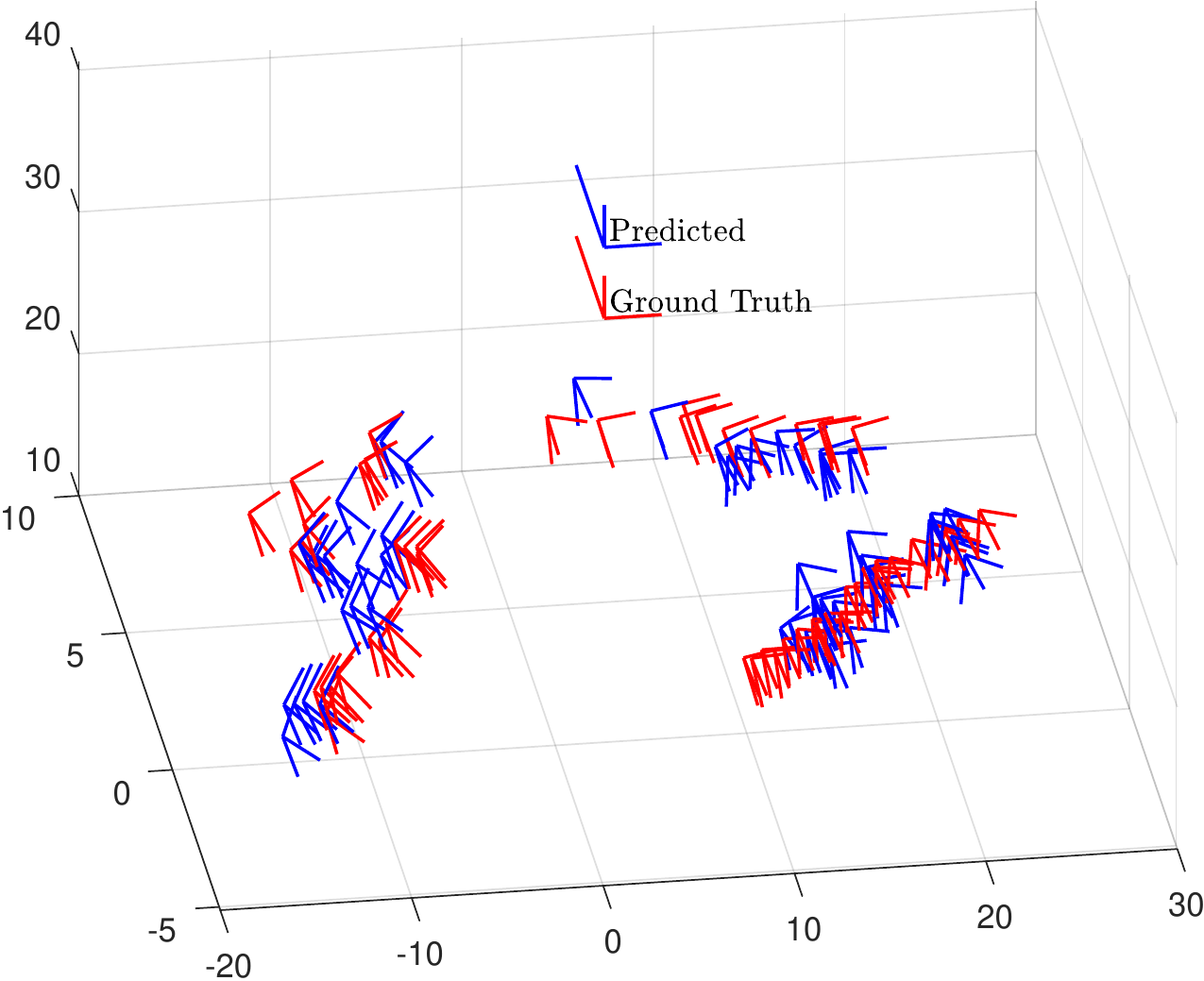}
        \includegraphics[width = 0.3\textwidth]{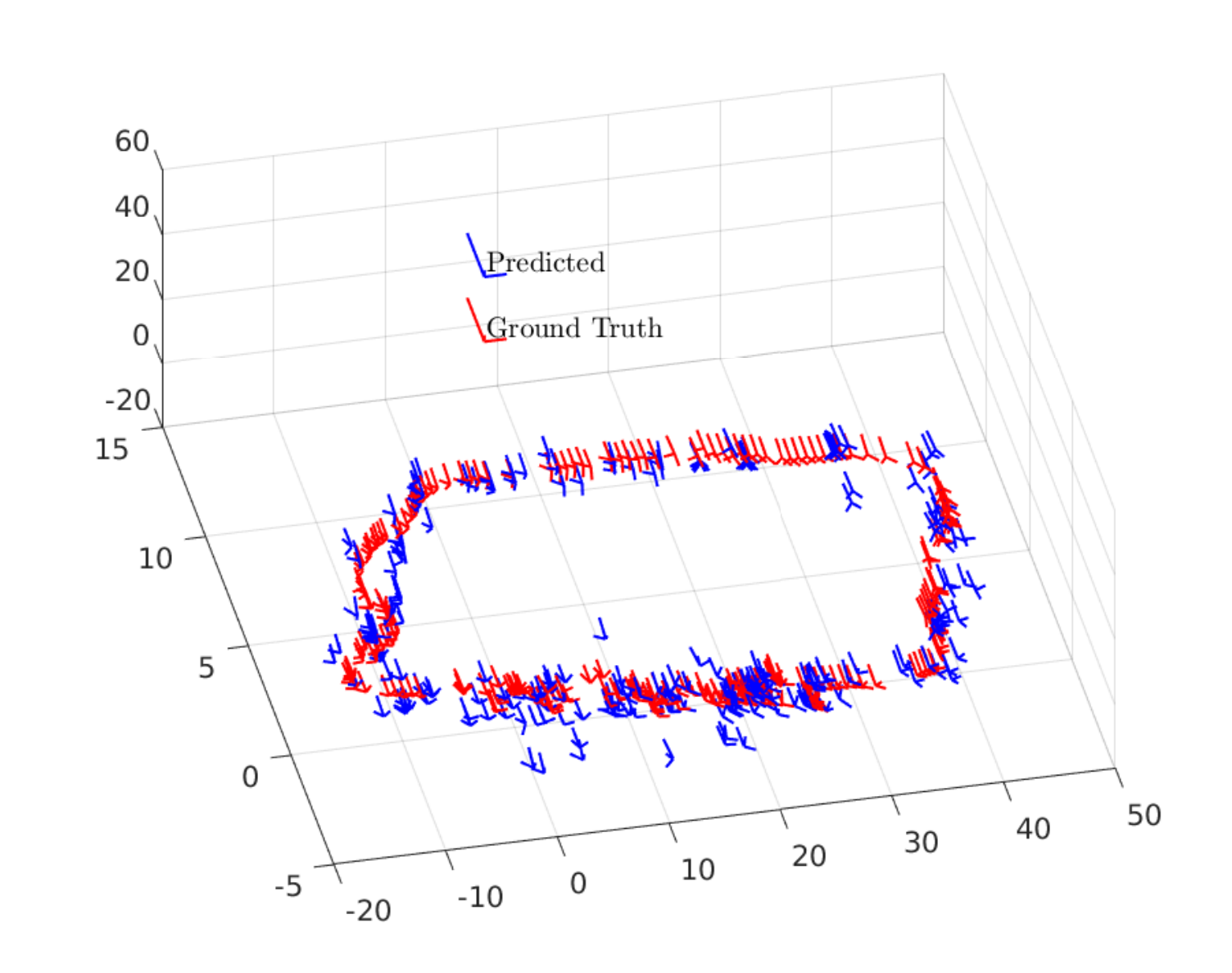}
        \caption{Cambridge Landmark dataset. From left to right: King's College, Old Hospital, StMary's Church}
    \end{subfigure}
    \caption{Qualitative results of SASSE (with PGO). The camera centers and their orientations predicted by SASSE (blue) and their corresponding ground truths (red) are shown (best viewed in color). }
\end{figure*}

\begin{figure}[h]
    \centering
    \includegraphics[width = 0.75\columnwidth]{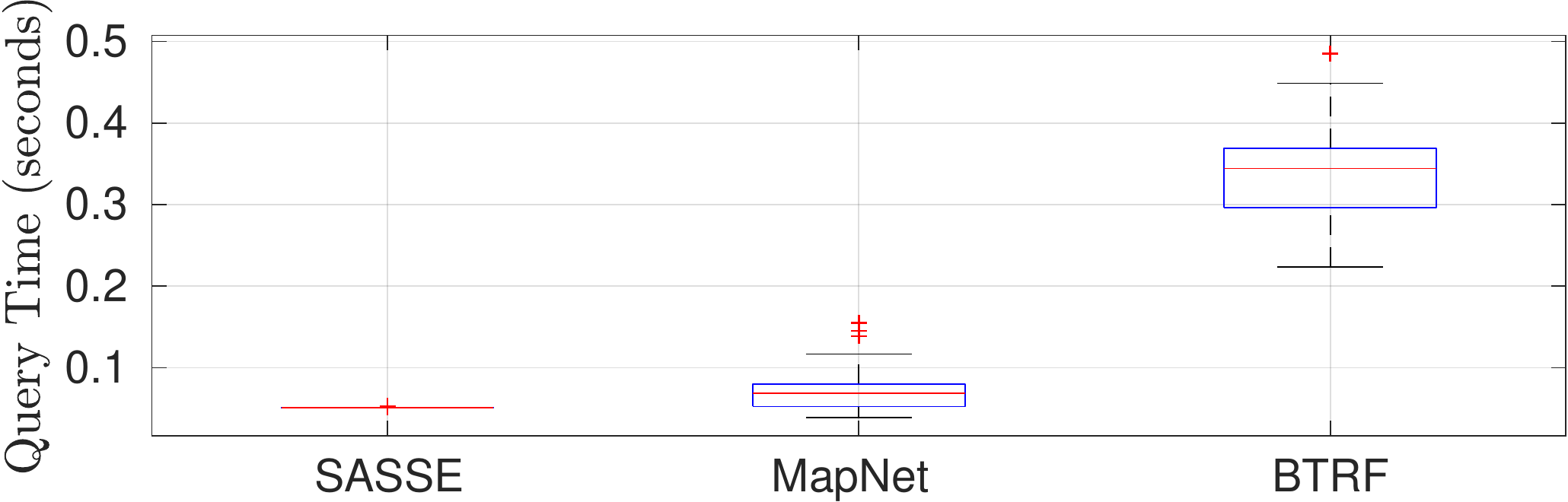}
    \caption{Box plots of query time for 1000 test images in the Heads sequence (7-Scenes dataset) without PGO. The query time of SASSE is significantly faster than the others.}
    \label{fig:query_time}
\end{figure}

\begin{table}[]
    \centering
    \resizebox{0.95\columnwidth}{!}{
        \begin{tabular}{@{}cccccc@{}}
            \toprule
                       & Chess         & Fire          & Heads          & Office & Pumpkin \\ \toprule
            SASSE    & \textbf{5.56} & \textbf{3.50} & \textbf{2.0}   & \textbf{7.65}        & \textbf{6.15}        \\
            BTRF~\cite{meng2017backtracking}       &    41.45    &   40.45  &   35.35     &    47.23     &     44.50  \\
            \bottomrule
        \end{tabular}
    }
    \caption{Training time (in minutes) for several sequences in the 7-Scene datasets. For SASSE, the time for feature extraction (VLAD) is also included. BTRF is executed based on the default parameters provided in the paper.}
    \label{table:training_time}
\end{table}

\begin{figure}[ht]
    \centering
    \begin{subfigure}{0.49\textwidth}        
        \centering
        \includegraphics[width=0.45\textwidth]{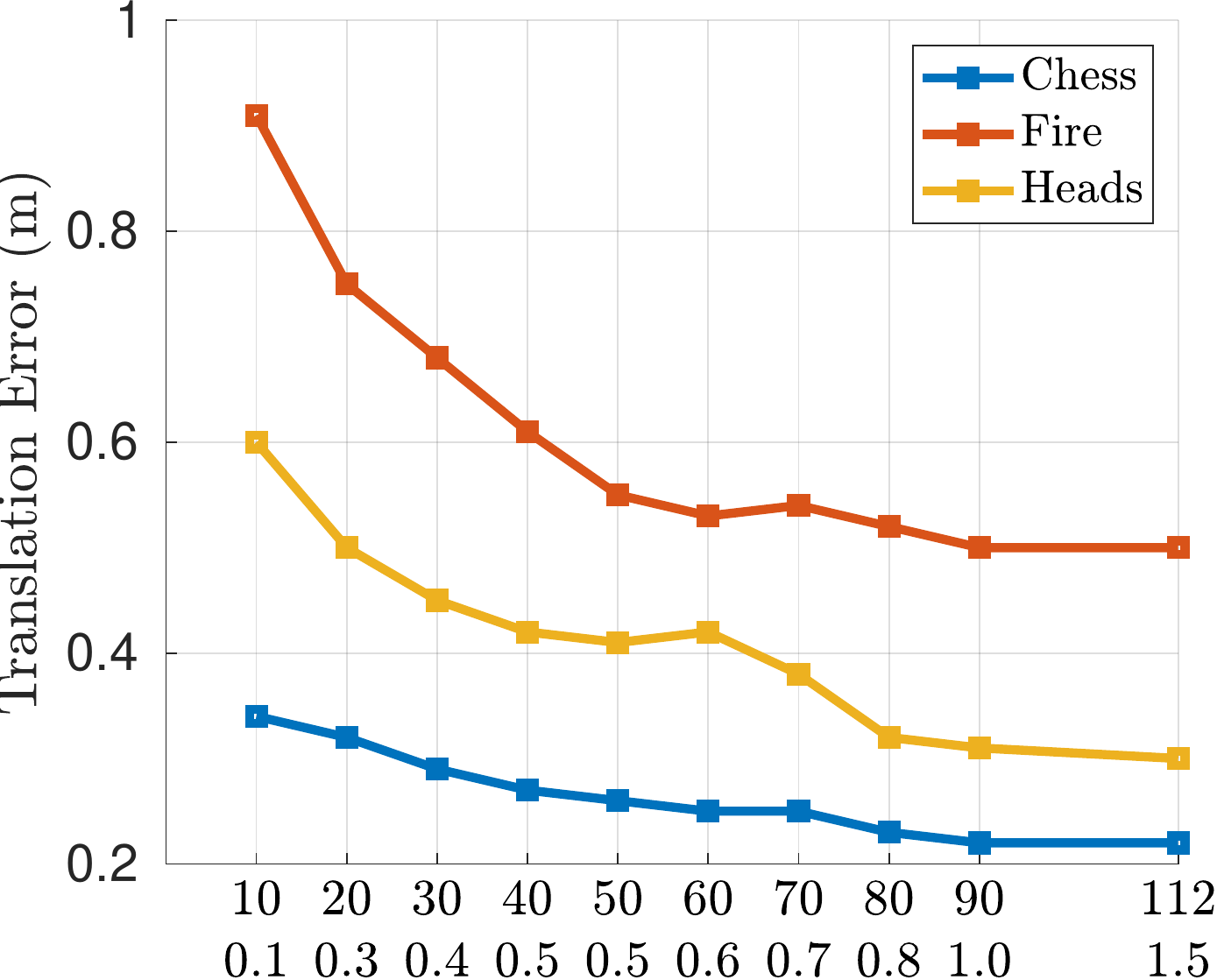}
        \includegraphics[width=0.45\textwidth]{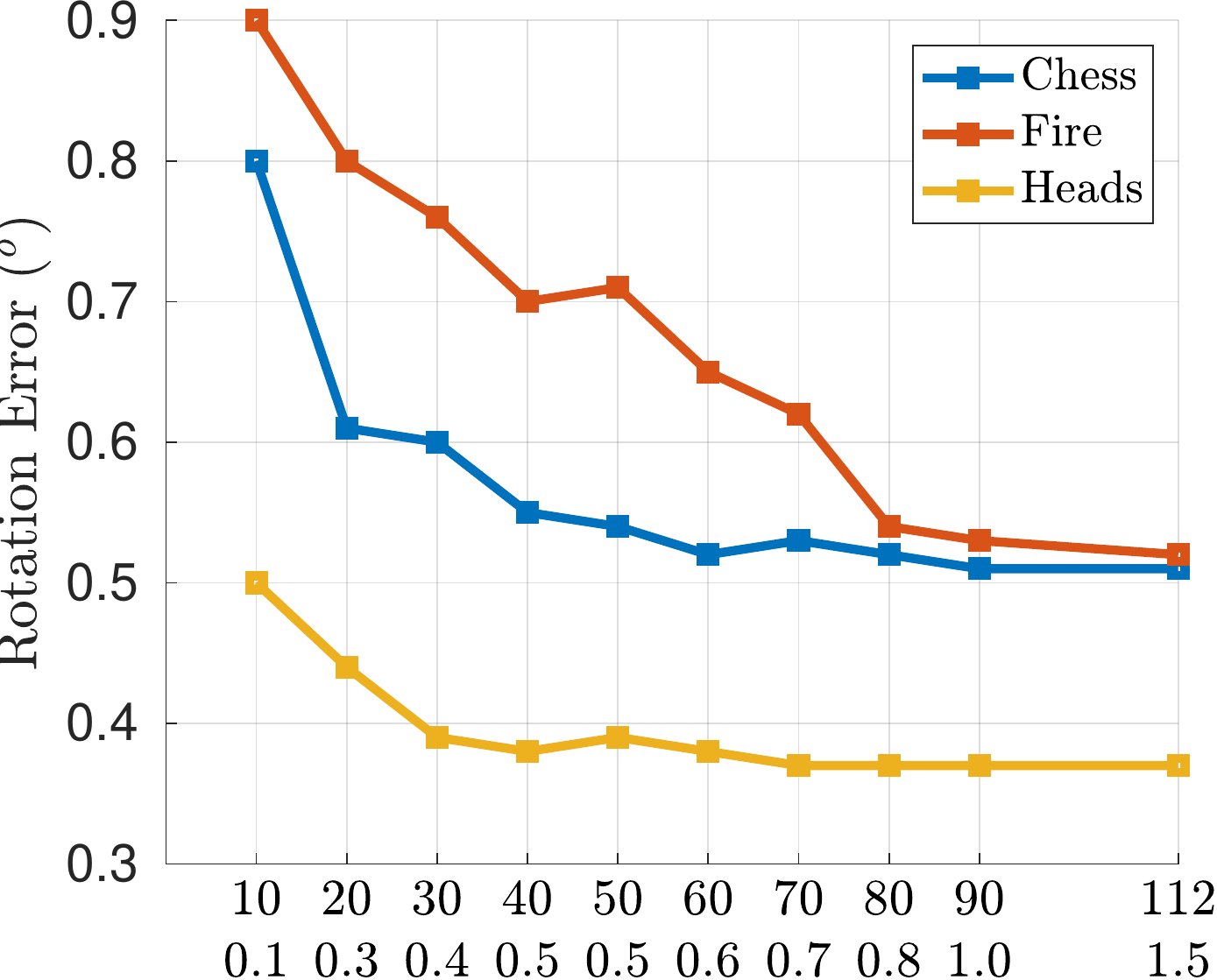}
        \caption{7-Scenes. The top row of x-axis shows $r$ and bottom row shows storage (in MB)}
        \label{subfig:trans_rot_vs_r_7scene}
    \end{subfigure}
    \begin{subfigure}{0.49\textwidth}        
        \centering
        \includegraphics[width=0.45\textwidth]{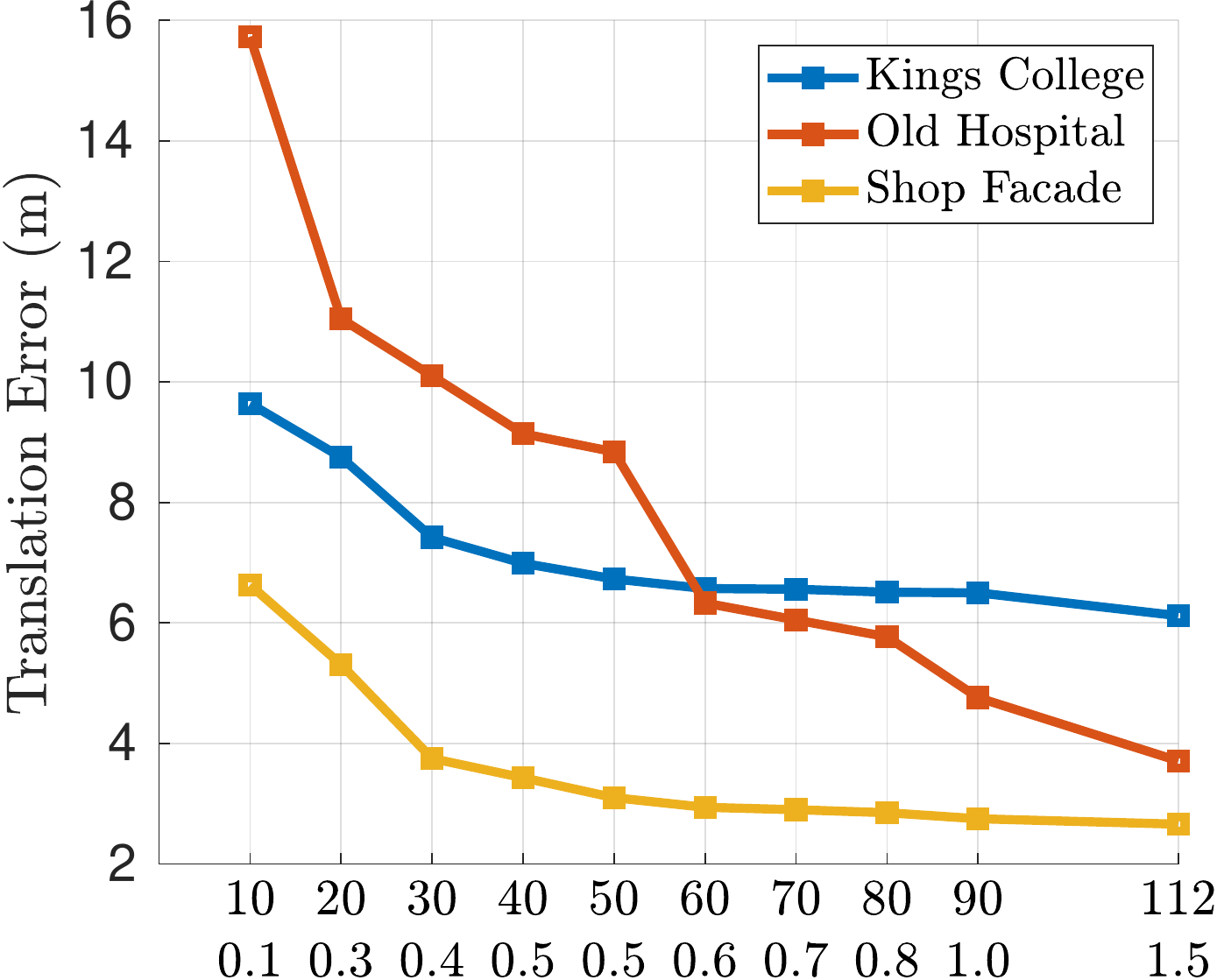}
        \includegraphics[width=0.45\textwidth]{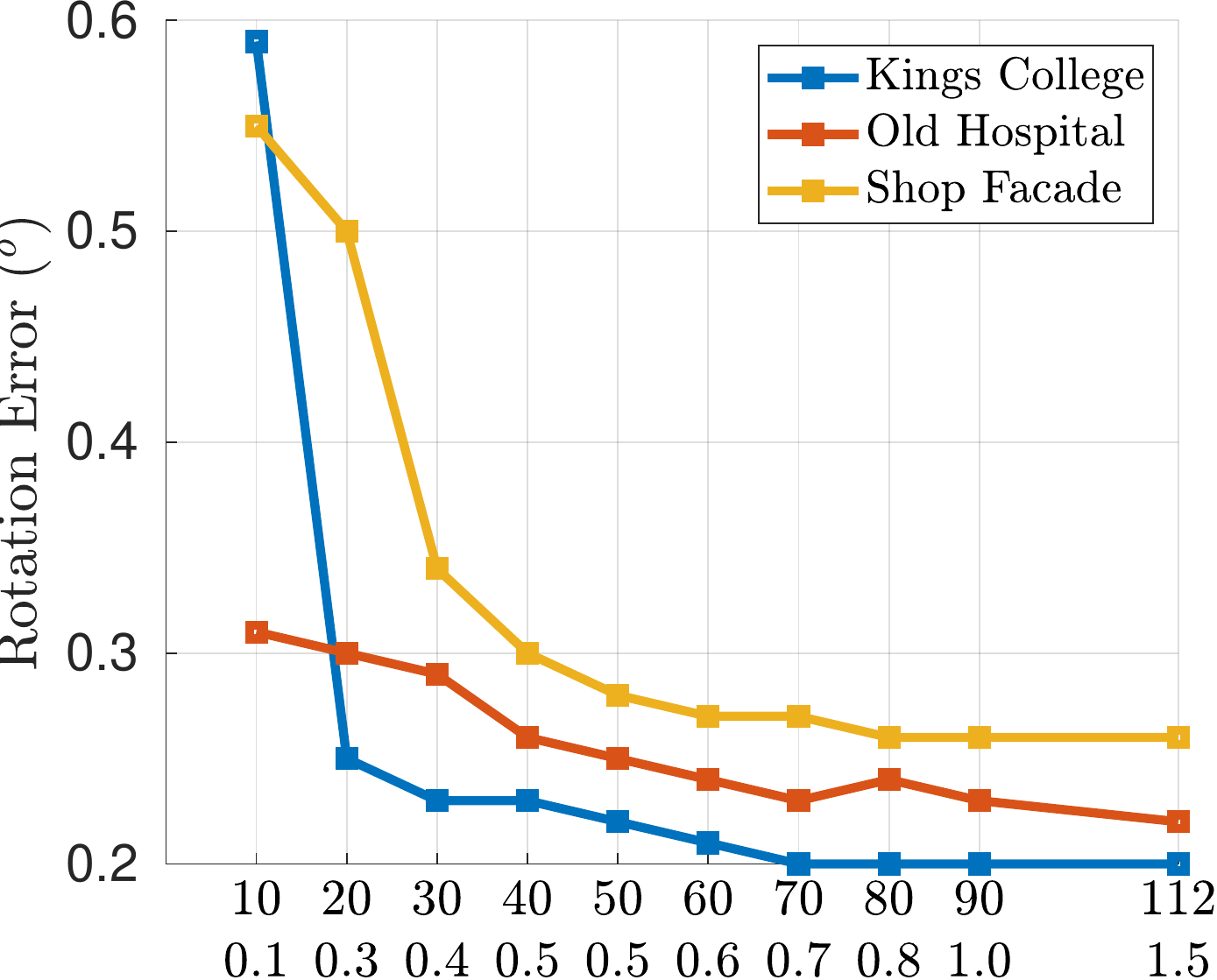}
        \caption{Cambridge Landmarks. The top row of x-axis shows $r$ and bottom row shows storage (in MB)}
        \label{subfig:trans_rot_vs_r_Cambridge}
    \end{subfigure}

    \caption{Median of translation and rotation error with increasing values of $r$ when the number of cluster is fixed to $1$.}
    \label{fig:trans_rot_vs_r}
\end{figure}

\begin{figure}[ht]
    \centering
    \begin{subfigure}{0.49\textwidth}        
        \centering
        \includegraphics[width=0.45\textwidth]{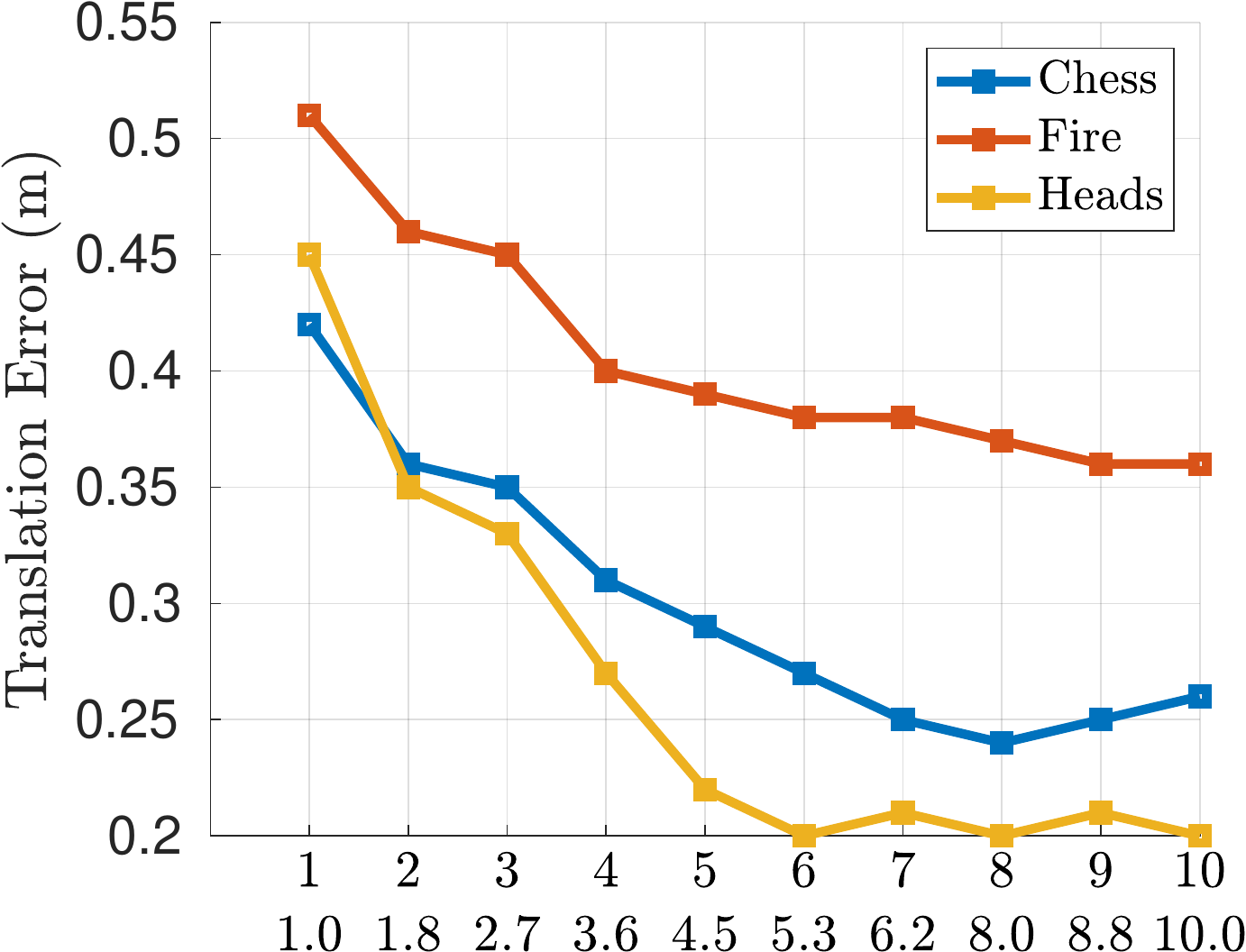}
        \includegraphics[width=0.45\textwidth]{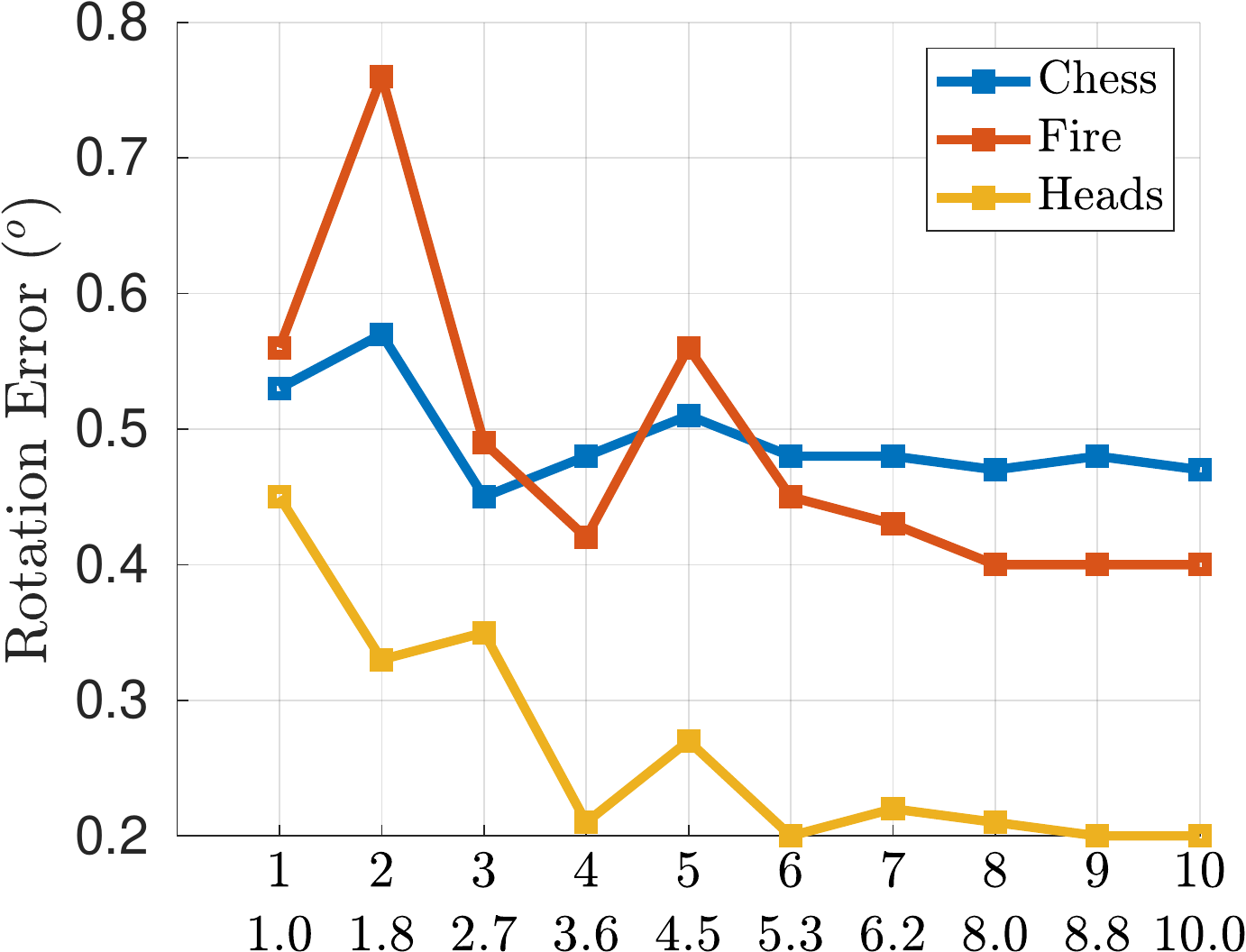}
        \caption{7-Scenes. The top row of x-axis shows $k$ and the second row shows storage (in MB).}
        \label{subfig:trans_rot_vs_k_7scene}
    \end{subfigure}
    \begin{subfigure}{0.49\textwidth}        
        \centering
        \includegraphics[width=0.45\textwidth]{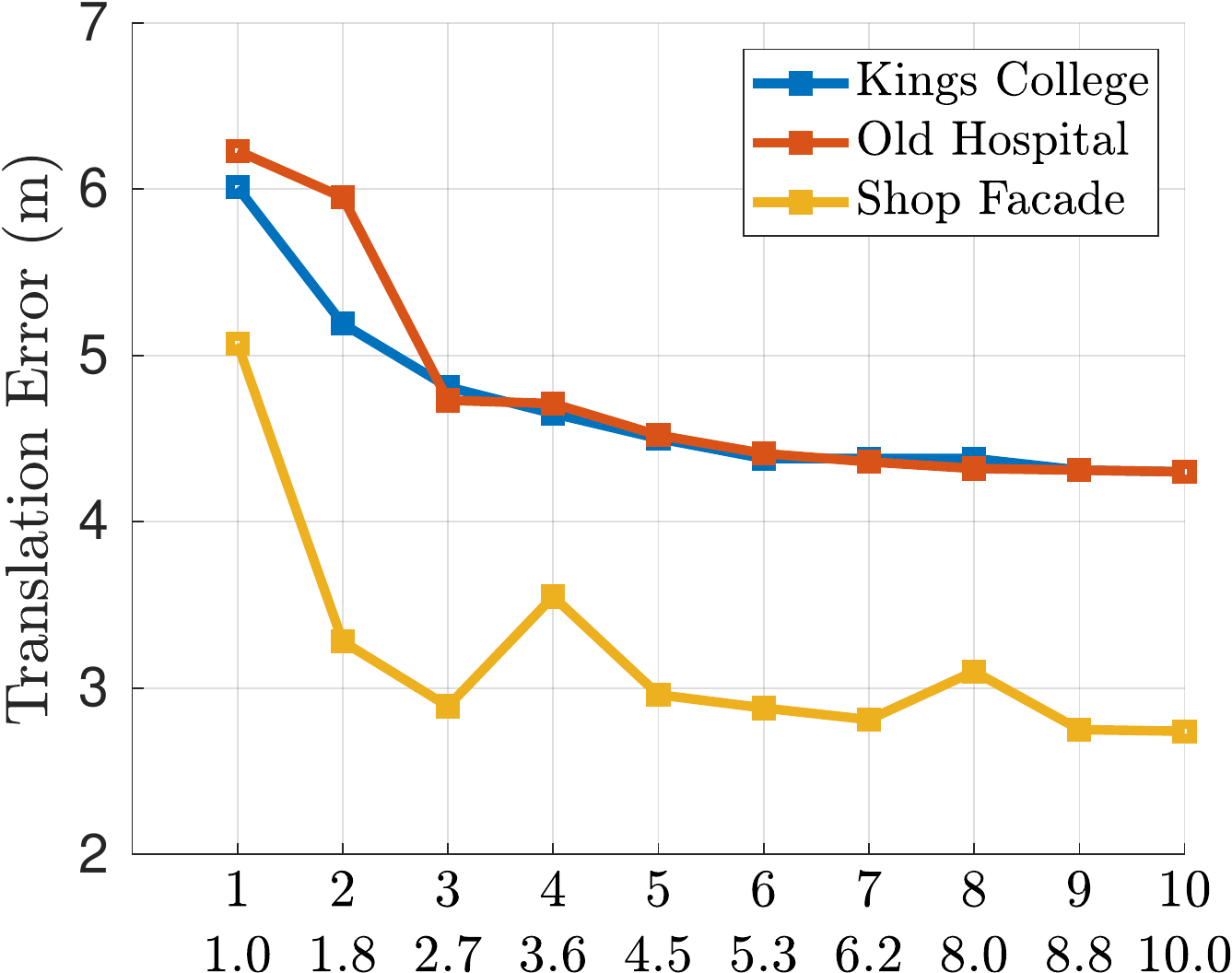}
        \includegraphics[width=0.45\textwidth]{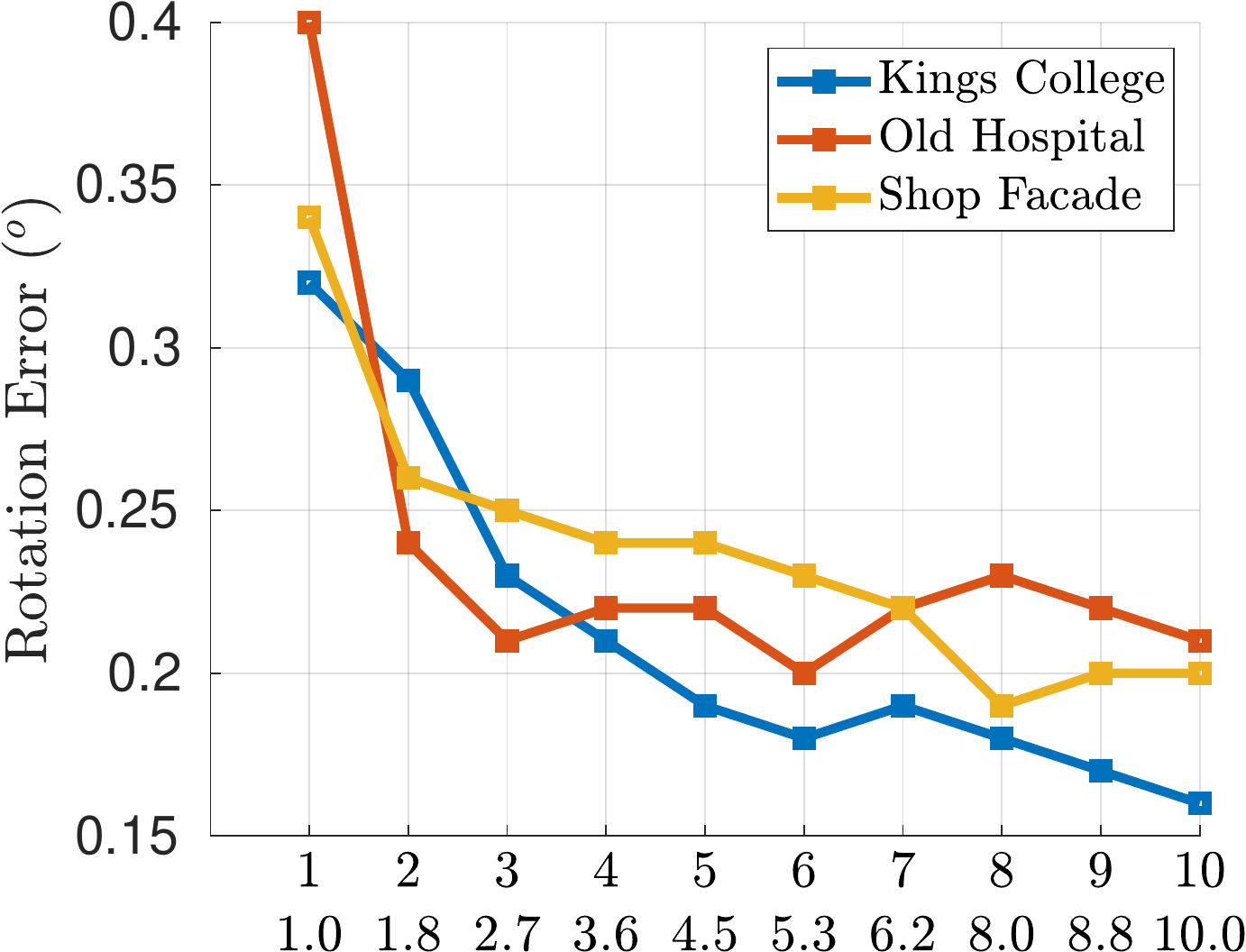}
        \caption{Cambridge Landmarks. The top row of x-axis shows $k$ and the second row shows storage (in MB).}
        \label{subfig:trans_rot_vs_k_Cambridge}
    \end{subfigure}
    \caption{Median of translation and rotation error with increasing values of $k$ when the embedding size $r$ is fixed to $50$.}
    \label{fig:trans_rot_vs_k}
\end{figure}

\begin{figure}[h]
    \centering
    \includegraphics[width = 0.49\columnwidth]{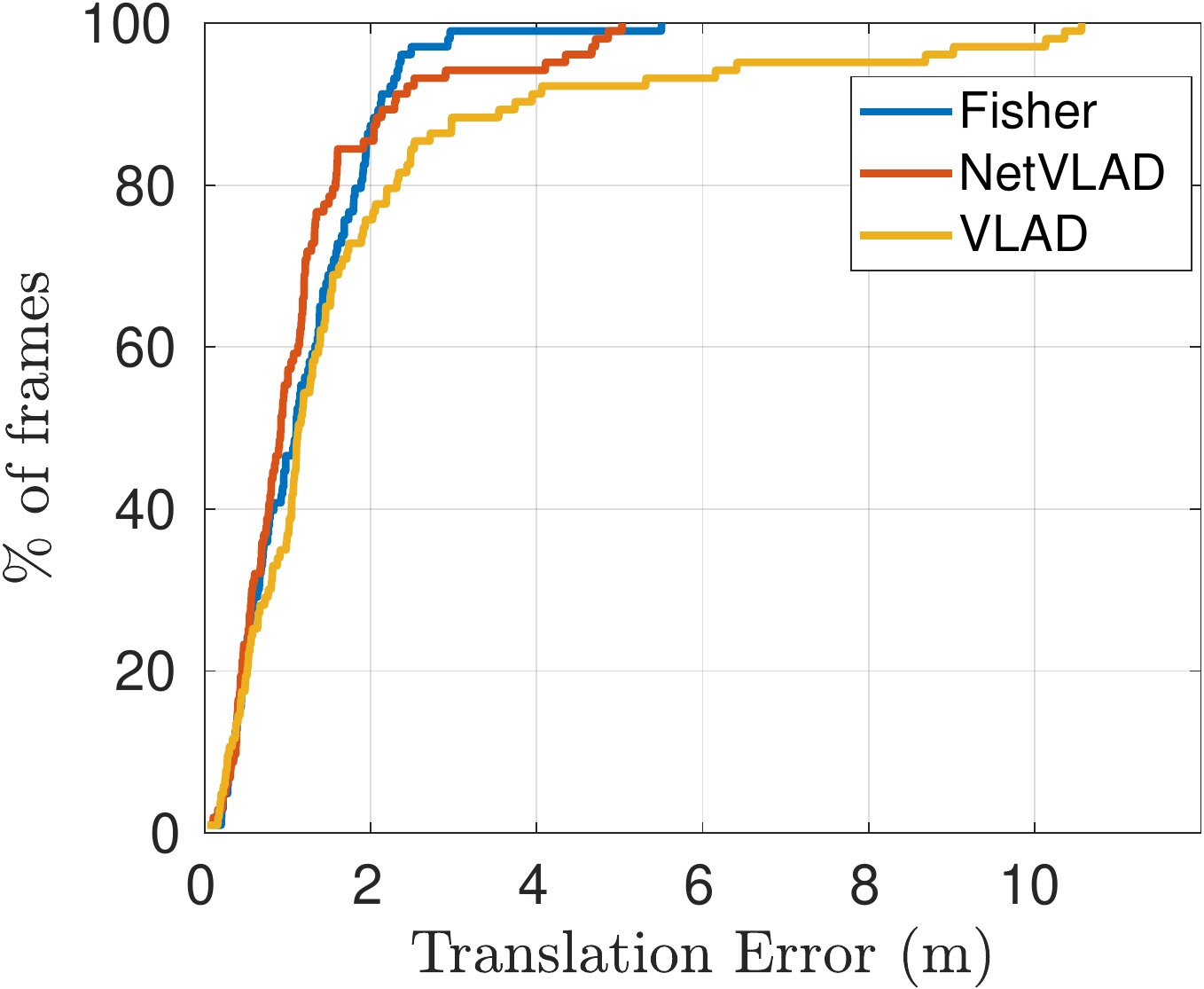}
    \includegraphics[width = 0.49\columnwidth]{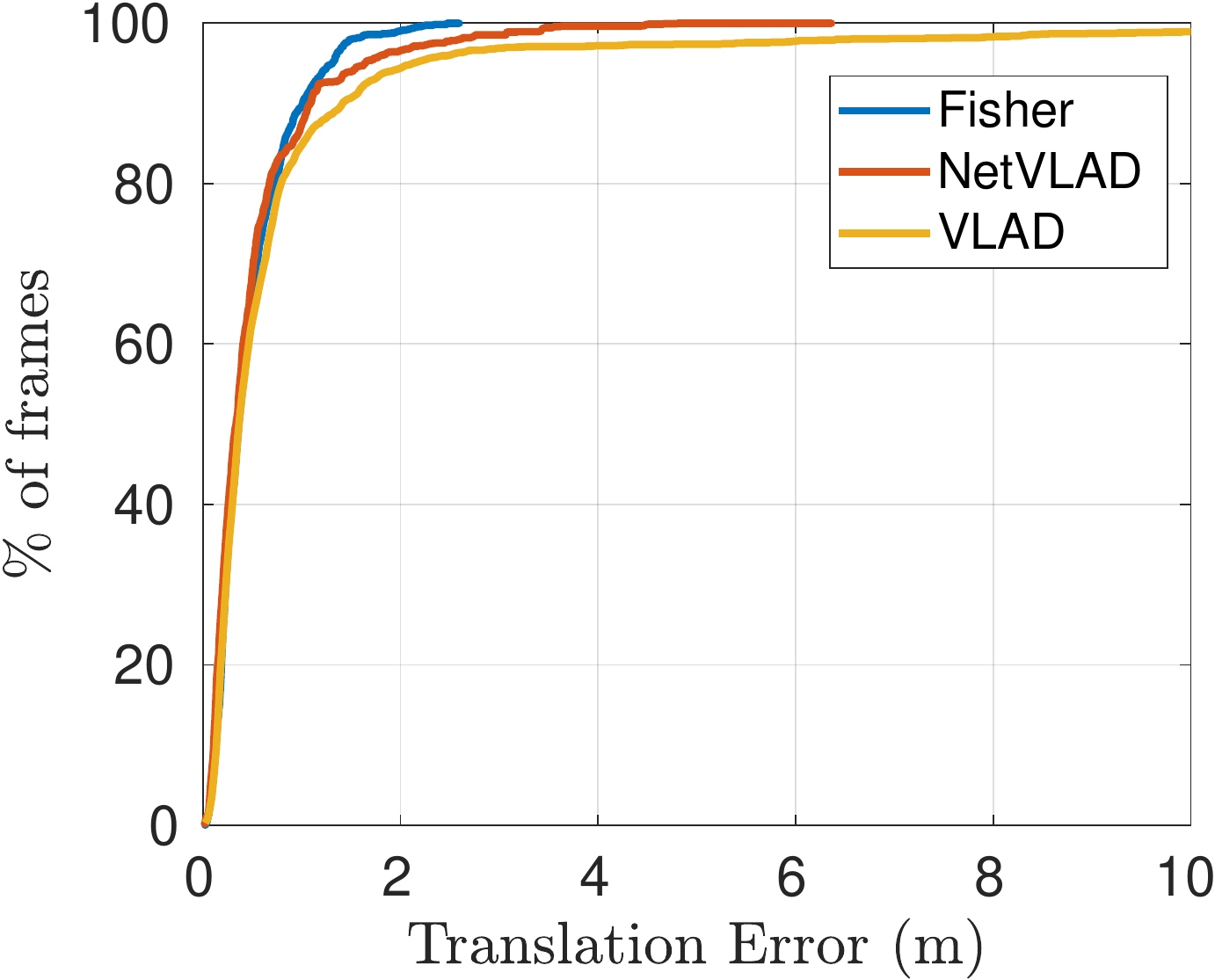}
    \caption{Cumulative translation error for three different image descriptors. \tb{Left}: Shop Facade. \tb{Right}: Chess.}
    \label{fig:cumm_error_features}
\end{figure}

\begin{figure}[h]
    \centering
    \includegraphics[width = 0.85\columnwidth]{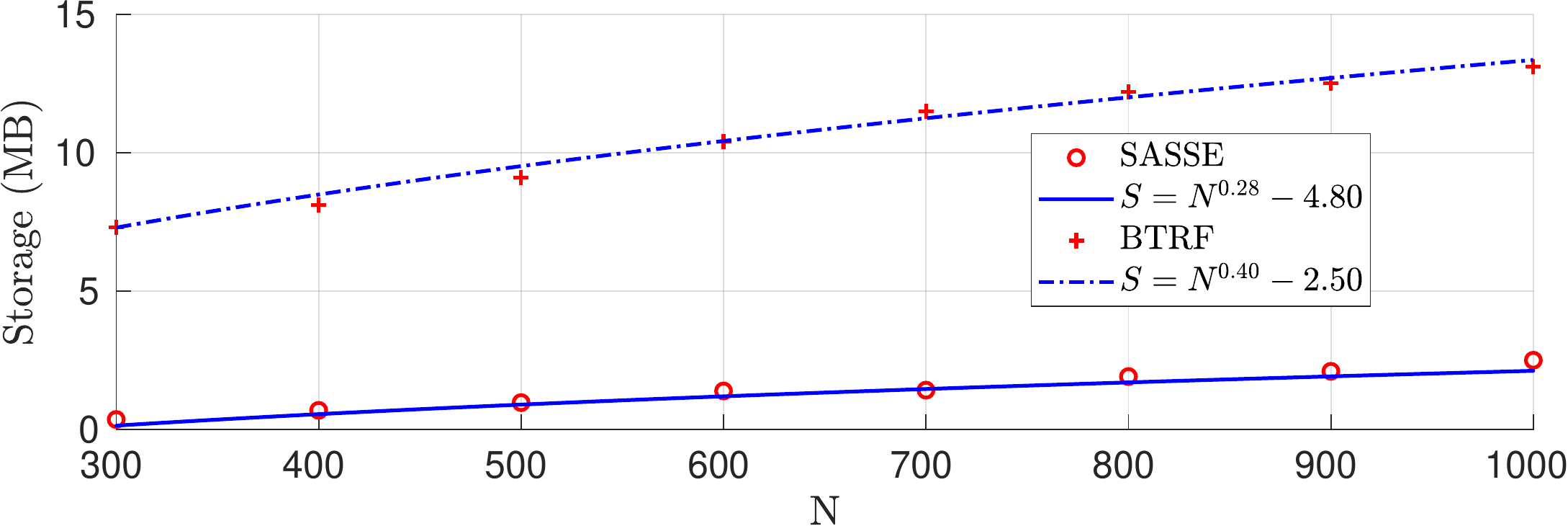}
    \caption{Scalability of SASSE and BTRF. X axis: Size of the training and testing data. Y axis: the amount of storage required to obtain translation errors of $\le 0.2m$ and and rotation errors of $\le 5.0\degree$.}
    \label{fig:scale}
\end{figure}

In this section, we conduct several experiments to evaluate the performance of our proposed method on several commonly used datasets. We also analyze the performance of SASSE under different parameter settings, and provide detailed information about the absolute storage required as well as its scalability.

\subsection{Datasets and Evaluation Metrics}
Two popular public datasets are used in our experiments: 7-Scenes Dataset~\cite{shotton2013scene}~\footnote{https://www.microsoft.com/en-us/research/project/rgb-d-dataset-7-scenes/}, which consists of RGB-D captures of seven different indoor scenes, and Cambridge Landmarks dataset~\cite{kendall2015posenet}, which contains video frames of 5 different scenes under large scale outdoor urban settings. 
Although depth information are provided in these datasets, we only make use of RGB images and their associated poses to train our system. 

To obtain global image descriptors for the training and testing images, we employ the VLFeat toolbox~\cite{vedaldi08vlfeat} to extract the VLAD~\cite{vlad} descriptors, where the number of codeword is set to $8$.  Unless stated otherwise, VLAD is used throughout all experiments. In Section~\ref{sec:feature_comparison}, the performance of our system with different types of feature will also be evaluated.

Our method is implemented in MATLAB R2018a. The source code of our method can be accessed at \url{http://tiny.cc/4ezt2y} and will be released with the publication of this paper. All experiments are executed on an Ubuntu Machine with an eight-core 4.20GHz CPU and 32GB of RAM.

Following other work on 6-DOF pose localization~\cite{kendall2015posenet,brahmbhatt2018geometry}, we report the median of translation and rotation errors (in meters and degree, respectively). More details on metrics for rotation can be referred to~\cite{huynh2009metrics}.

\subsection{Localization Accuracy and Required Storage}
In order to demonstrate the localization ability of our framework, we first report the median of translation and rotation errors on the 7-Scenes and Cambridge Landmark datasets, together with theirs required storages. A detailed study on the effect of parameters will be presented in latter sections. 

Table~\ref{table:results_7scenes} and Table~\ref{table:results_Cambridge} shows the results of SASSE and other methods on the two benchmarked datasets. Observe that the translation errors obtained by our method is very competitive compared to MapNet and several PoseNet's variants. Note that the results of SCoRe Forest~\cite{shotton2013scene} is trained with RGB-D data. Notably, the rotation provided by our method outperforms the results from other methods, which demonstrates the ability to accurately regress poses of our system. For most of the sequences, the rotation error achieved by our system is less than $1\degree$, while the errors of PoseNet and MapNet are much larger.

Table~\ref{table:storage} shows the absolute storage required by our method to obtain the results shown in Table~\ref{table:results_7scenes} and Table~\ref{table:results_Cambridge}. For SASSE, the storage of VLAD's codebook (for feature extraction during inference) is also included. Our absolute storage footprint is significantly smaller  compared to PoseNet~\cite{kendall2015posenet}, MapNet~\cite{brahmbhatt2018geometry}, and tree-based approaches such as BTRF~\cite{meng2017backtracking} which are among the state-of-the-art methods for 6-DOF pose regression. Note that in contrast to the fixed network structures of PoseNet or MapNet, SASSE allows the flexible configuration of the training process. Therefore, we are able achieve much smaller absolute storage while maintaining a competitive performance.

\subsection{Testing and Training Time}
Besides the ability to encode a large amount of training data into a compact set of parameters with  very small storage footprint, our design allows the testing time to be significantly shorter than current approaches. This is shown in Figure~\ref{fig:query_time}, where we plot the query time of SASSE in comparison with MapNet~\cite{brahmbhatt2018geometry} and BTRF~\cite{meng2017backtracking}. As can be seen in this figure, SASSE runs significantly faster than MapNet and BTRF due to the fact that the pose inference in SASSE requires only simple matrix operations. On the other hand, in order to achieve good results, tree-based methods such as BTRF requires an additional step of performing 2D-3D matching to obtain the pose leading to the increase in total run time. Note that we execute MapNet on GPU to report the time Figure~\ref{fig:query_time}, while our method only runs on CPU with a MATLAB implementation. 

Moreover, the time required to train our system is also faster than other methods. Table~\ref{table:training_time} reports the training time of our system for several sample sequences in the 7-Scenes dataset. Note that we have also included the time needed to extract the VLAD descriptors. Observe that SASSE requires less than 10 minutes for all the datasets, while the training time for BTRF is around 45 minutes. It is also well-known that the training time for deep-network based approaches such as PoseNet requires much higher training time, thus we omit them from Table~\ref{table:training_time}.

\subsection{Performance under Different Parameter Settings}
In this section, we study the performance of the system under different parameter settings. We focus on the two main parameters that can be adjusted to achieve the desired storage, namely, the embedding size $r$ and the number of clusters $k$.
\subsubsection{Embedding Size $r$}
\label{sec:embedding_size_tuning}
In Figure~\ref{fig:trans_rot_vs_r}, we plot the translation and rotation error of SASSE under different values of $r$ with the number of cluster $k$ is fixed to $1$. Note that the values of required storage are  also displayed on the bottom row of x-axis. As can be seen, SASSE performs well even for small values of $r$, demonstrating the effectiveness of the embedding step. When $r$ increases, the performance becomes even better, as more information is retained during the training process. This behavior can be observed for all sequences in the two datasets. 
\subsubsection{Number of Clusters $k$}
\label{sec:cluster_size_tuning}
Next, we study the effect of the clustering in a similar manner. Figure~\ref{fig:trans_rot_vs_k} plots the performance of SASSE across several values of $k$ with a fixed value of $r=50$. When the number of clusters increases the performance improves. However, when we continue to increase $k$, the decrease in errors is no longer significant. This shows that the good performance of SASSE relies mainly on the learning algorithm discussed in Section~\ref{sec:training}, while the clustering step is for the purpose of preventing under-fitting.

\subsection{Performance under different types of image descriptor}
\label{sec:feature_comparison}
SASSE, as discussed, can be trained with different types of image descriptors. Figure~\ref{fig:cumm_error_features} plots the cumulative translation errors for two sequences (King's College and Chess) with three commonly used descriptors: VLAD~\cite{vlad}, NetVLAD~\cite{netvlad} and Fisher~\cite{perronnin2007fisher}, where it can be observed that the performances of SASSE with these descriptors are comparable. Although the CNN-based descriptor NetVLAD shows a slight increase in performance, the storage of its network for feature extraction during inference will also increase the absolute storage. On the other hand, traditional hand-crafted features can also provide competitive results with much less storage. 

\subsection{Storage Scalability}
To test the scalability of SASSE, we investigate the required storage when the number of training data increases. We extract $N$ training and $N$ testing data points from the Heads sequence, with $N$ increases from $300$ to $1000$. For each value of $N$, we run the SASSE with different sets of parameters and report the smallest amount of storage required such that the median of translation errors is less than $0.2m$ and the median of rotation errors is less than $5.0\degree$. The storage is compared against BTRF~\cite{meng2017backtracking}. To adjust the required storage for BTRF, we change the number of regression trees and the number of levels in each tree. Figure~\ref{fig:scale} plots the results. To analyze the dependence of total storage $S$ on database size $N$, we fit a function of the form. $S = N^{a} + b $ to the data points in Figure~\ref{fig:scale}. The plots of these functions are also shown, where the $(a,b)$ tuples for SASSE and BTRF are $(0.28, -4.8)$ and $(0.4, -2.5)$, with the mean squared errors (MSE) of 0.05 and 0.07, respectively.

Observe that both SASSE and BTRF show the tendency to scale sub-linearly with the number of training data. However, since $S_{\text{SASSE}} = O(N^{0.28})$, while $S_{\text{BTRF}} = O(N^{0.4})$, it is evident that SASSE achieves better scalability in terms of required storage due to our efficient encoding method. BTRF, on the other hand, needs to store feature information at each leaf node, requiring larger storage footprint. In our experiments, at least $3$ trees are required, while the number of levels in each tree must be greater than $8$ in order for BTRF to achieve good localization results. 



\section{Conclusion and Future Work}

In this paper, we have presented a novel storage-efficient algorithm for 6-DOF localization (SASSE). While providing competitive localization results compared to existing state-of-the-art approaches, our method has several additional advantages: it requires a much smaller storage footprint, has a more significantly sub-linear storage scaling profile, is agnostic to image descriptor type and hence can easily utilize descriptors suited to the specific application without re-training or learning, achieves faster training and deployment speeds, and is easily adapted to suit a wide range of computational resource scenarios.

In the future work, SASSE could be extended for applications where 6-DOF poses are not readily available. In such scenarios, one can consider the pose contains only 1-DOF single translation (to indicate to current location) and the same encoding technique as SASSE can be applied. Furthermore, the learning of binary labels can be improved by employing recent developments in multiple-label classification research such as tree learning approaches~\cite{prabhu2014fastxml} or improving the embedding techniques. Additionally, besides being used for localization, the techniques underlying the SASSE system can also potentially be utilized for several similar 6-DOF estimation tasks in robotics, such as object pose estimation for robot grasping and manipulation.

\bibliographystyle{plainnat}
\bibliography{sublinear}

\end{document}